\documentclass[journal]{IEEEtran}

\usepackage[dvipsnames]{xcolor}
\usepackage{adjustbox}
\usepackage{booktabs}
\usepackage{threeparttable}
\usepackage{caption}
\usepackage{subcaption}
\usepackage{amssymb}
\usepackage[colorlinks]{hyperref}
\usepackage{hyperref}
\usepackage{amsmath}
\usepackage{colortbl}
\usepackage{hhline}
\usepackage{multicol}
\usepackage{chngcntr}
\usepackage{mathrsfs}
\usepackage{mathtools}
\usepackage{cite}
\usepackage{graphicx}
\usepackage{multirow}
\usepackage{rotating}
\hypersetup{
  colorlinks   = true,
  citecolor    = blue,
  linkcolor    = blue,
  urlcolor     = blue,
}

\usepackage{amsthm}
\usepackage{cancel}
\usepackage{ragged2e}
\usepackage[switch, modulo]{lineno} 
\usepackage[shortlabels]{enumitem}
\usepackage{gensymb}

\def\etal{\textit{et al.}}
\def\ie{\textit{i.e.}}
\def\eg{\textit{e.g.}}

\begin{document}
\bstctlcite{IEEEexample:BSTcontrol}

\title{A Teleoperation System with \\Impedance Control and Disturbance Observer \\for Robot-Assisted Rehabilitation} 

\author{Teng~Li,~\IEEEmembership{Graduate Student Member,~IEEE}

\thanks{This research is supported in part by the Canada Foundation for Innovation (CFI) under grants LOF 28241 and JELF 35916, in part by the Government of Alberta under grants IAE RCP-12-021 and EDT RCP-17-019-SEG, in part by the Government of Alberta's grant to Centre for Autonomous Systems in Strengthening Future Communities (RCP-19-001-MIF), in part by the Natural Sciences and Engineering Research Council (NSERC) of Canada under grants RTI-2018-00681, RGPIN-2019-04662, and RGPAS-2019-00106, and in part by the Edmonton Civic Employee Charitable Assistance Fund.}

\thanks{T. Li is with the Department of Electrical and Computer Engineering, Faculty of Engineering, University of Alberta, Edmonton T6G 1H9, Alberta, Canada. E-mail: {\tt\small teng4@ualberta.ca}}

}

\markboth{} 
{Shell \MakeLowercase{\textit{et al.}}: Bare Demo of IEEEtran.cls for IEEE Journals}

\maketitle
\thispagestyle{plain}
\pagestyle{plain}

\begin{abstract}
Physical movement therapy is a crucial method of rehabilitation aimed at reinstating mobility among patients facing motor dysfunction due to neurological conditions or accidents. Such therapy is usually featured as patient-specific, repetitive, and labor-intensive. The conventional method, where therapists collaborate with patients to conduct repetitive physical training, proves strenuous due to these characteristics. The concept of robot-assisted rehabilitation, assisting therapists with robotic systems, has gained substantial popularity. However, building such systems presents challenges, such as diverse task demands, uncertainties in dynamic models, and safety issues. To address these concerns, in this paper, we proposed a bilateral teleoperation system for rehabilitation. The control scheme of the system is designed as an integrated framework of impedance control and disturbance observer where the former can ensure compliant human-robot interaction without the need for force sensors while the latter can compensate for dynamic uncertainties when only a roughly identified dynamic model is available. Furthermore, the scheme allows free switching between tracking tasks and physical human-robot interaction ($p$HRI). The presented system can execute a wide array of pre-defined trajectories with varying patterns, adaptable to diverse needs. Moreover, the system can capture therapists' demonstrations, replaying them as many times as necessary. The effectiveness of the teleoperation system is experimentally evaluated and demonstrated.
\end{abstract}


\IEEEpeerreviewmaketitle

\section{Introduction}

Paralysis is a side effect of stroke that requires physical therapy for rehabilitation, without which the stroke survivor may lose motor control on some body parts in the long term. Patients with paralysis suffer from an inability to move some body parts, reduced mobility, restricted range of motion, and muscle stiffness \cite{najafi2020usingPotentialFieldTMECH}. During conventional physical therapy, the therapist needs to help the patient to move the affected body part, \eg{}, arm, repetitively to regain mobility \cite{najafi2020usingPotentialFieldTMECH}. This repetitive movement therapy training is labor-intensive and strenuous for the therapist but necessary for the patient. Beyond stroke, patients with motor dysfunction due to other conditions (\eg{}, cerebral palsy, Parkinson's disease) or injuries on body parts (\eg{}, fall, car accident), may also need physical therapy which largely increases its demand. Robot-assisted rehabilitation is a promising solution where a robot can be deployed to assist the therapist in helping the patient perform movement therapy practices repetitively with a certain degree of accuracy \cite{ocampo2019visualHapticColocation2dAR, najafi2020usingPotentialFieldTMECH}.

There is an abundance of work that has been done for robot-assisted rehabilitation over the past three decades while many of them involved only one robot manipulator \cite{hogan1995interactiveUSPatent, riener2005robotAidedUpperLimb, meng2015recentLowerLimbRehab, sharifi2020assistAsNeededCEP}. Some new techniques can be integrated into the system to enhance its capability. For example, Ocampo and Tavakoli built a visual-haptic colocation system for robot-assisted rehabilitation exercises by involving a 2D augmented reality (AR) display \cite{ocampo2019visualHapticColocation2dAR}. By providing spatial AR visual feedback together with haptic feedback, the system enabled better task performance in terms of task completion time.

Learning from demonstration (LfD) is a popular option for the robot to learn and help with repetitive rehabilitative and assistive practices. Najafi \etal{} proposed a framework using a potential field function with a velocity field controller to learn and produce the therapist's assistance in robot-assisted rehabilitation where the therapist just needs to demonstrate once \cite{najafi2020usingPotentialFieldTMECH}. The potential field function is used to model and reproduce the therapist's motion and assistive force, while the velocity field controller is designed to compensate for and regulate the patient's deviation.

In recent years, teleoperation systems have gained increasing interest due to the additional advantages of allowing in-home environment therapy and avoiding physical contact for better hygienic conditions. Sharifi \etal{} proposed an approach for therapist-patient collaboration in a telerobotic system with time delays \cite{sharifi2018patientJMR}. The approach refers to a bilateral impedance control strategy that can filter out the patient's involuntary hand tremor. Their telerobotic system allows the patients to move their limbs on the master robot side while being assisted as much as needed by the online assistive force provided by the therapist on the second robot side. Later, Sharifi \etal{} employed a similar strategy for assist-as-needed tele-rehabilitation \cite{sharifi2020assistAsNeededCEP}. With this strategy, the system can minimize the therapist's movements while providing a better perception to the therapist about the patient's issues and recovery status in motor control ability. As an additional benefit of the strategy, the patient can receive assistance from two sources, \ie{}, the adjustable impedance model, and the exerted force by the therapist.

Safety and compliance are the paramount concerns when developing a robotic system for rehabilitation, either for collaboration or teleoperation. Impedance control is a commonly adopted control scheme considering that it can provide compliant robot behavior thus ensuring safety. Impedance control has its intrinsic property of compliance which makes it more suitable for human-robot interaction \cite{song2019tutorialSurveyImpedanceControl, teng4IROS2022NDOB}. A robot with an impedance controller can be programmed to be soft (compliant) or rigid (non-compliant) as necessary, which ensures a safe human-robot interaction \cite{bruno2010roboticsBookSiciliano}. Moreover, the measurement of the interaction force is not needed for impedance control. In brief, the sensor-free and compliance properties make impedance control a popular choice for human-robot interaction scenarios. Another common problem in building such a robotic system is that the dynamics of the robot might be unknown or only a roughly identified dynamic model be available. An integrated framework of combining impedance control with disturbance observer has been proposed in our previous work \cite{teng4IROS2022NDOB}. By involving a disturbance observer to compensate for the modeling error and other uncertainties, an accurate impedance control can be achieved even with an inaccurate dynamic model. 

Due to the diverse motor inabilities among the patients, each individual patient requires individualized physical therapy for rehabilitation. Even for the same patient, different therapy programs are also required at different rehabilitation phases. For example, a patient may not be able to move the limb at all at the very beginning, which means that passive movement is needed with the help of the therapist. Gradually, the patient can voluntarily move the limb in a small area, and in this phase, assist-as-needed rehabilitation may be a better option \cite{sharifi2020assistAsNeededCEP}. When the patient is able to move the limb around relatively freely, more complicated moving trajectories or tasks can be designed for the patient to further enhance the rehabilitation.

To better meet these requirements, a bilateral teleoperation system is proposed for robot-assisted rehabilitation in this paper, which integrates several features, including pre-defined trajectory tracking, LfD, and force feedback, into the same teleoperation system under the same control architecture. The designed control architecture, a main contribution of this work, enables free switching between pre-defined trajectory tracking mode and physical human-robot interaction ($p$HRI) mode. With the control architecture, the proposed teleoperation system has the following detailed features, 

\begin{enumerate}[label={(\arabic*)}]
  \item Force feedback is provided on the master robot (therapist) side. With this feature, the therapist can better understand the status of the patient on the second robot side.
  \item A large variety of pre-defined trajectories with different patterns can be programmed thus the therapist can select the ones that are most suitable to individual patients.
  \item The therapist on the master robot side can design customized trajectories for individual patients as needed.
  \item Similar to the concept of LfD, the customized trajectory demonstrated by the therapist can be recorded, and then it can be replayed by the system on its own as many times as needed without losing reproduction fidelity.
\end{enumerate}

The remainder of this paper is organized as follows: Section \ref{methods} is devoted to introducing the general methods including dynamics, impedance control, and disturbance observer. Section \ref{robots} presents the details of the proposed teleoperation system, including the master/second robot kinematics, dynamics, and force feedback rendering. Section \ref{experiments} presents experiments and corresponding results for evaluating the system. Section \ref{conclusions} gives the concluding remarks.

\section{Methods}
\label{methods}

\subsection{Robot Dynamics and Impedance Control}

A general dynamic model for an $n$-degree-of-freedom (DOF) rigid robot \cite{fong2019} can be given by
\begin{equation}
 \label{eqn_dynModel_general}
    \mathbf{ \underbrace{\mathbf{M(q)}}_{\hat{M}+\varDelta M} \ddot{q} + \underbrace{\mathbf{S(q,\dot{q})}}_{\hat{S}+\varDelta S} \dot{q} + \underbrace{\mathbf{g(q)}}_{\hat{g}+\varDelta g} + \boldsymbol{\tau}_{fric}(\dot{q}) = \boldsymbol{\tau} + }
    \underbrace{\mathbf{\boldsymbol{\tau}_{ext}}}_\mathbf{J^T F_{ext}} 
\end{equation}
\noindent
where $\mathbf{q, \dot{q}, \ddot{q}}$ are the actual joint angle, velocity, and acceleration, respectively, $\mathbf{M} \in \mathbb{R}^{n \times n}$ denotes the inherent inertia matrix, $\mathbf{S}\in \mathbb{R}^{n \times n}$ denotes a matrix of the Coriolis and centrifugal forces, $\mathbf{g}\in \mathbb{R}^{n}$ represents the gravity vector. $\mathbf{\hat{M}, \hat{S}, \hat{g}}$ represent their estimates, while $\mathbf{\varDelta M, \varDelta S, \varDelta g}$ are the corresponding estimated errors. $\mathbf{\boldsymbol{\tau}_{fric}} \in \mathbb{R}^{n}$ is joint friction,
$\mathbf{\boldsymbol{\tau}} \in \mathbb{R}^{n}$ is the commanded joint torque vector, $\mathbf{\boldsymbol{\tau}_{ext}} \in \mathbb{R}^{n}$ is the torque caused by the external force, $\mathbf{F_{ext}} \in \mathbb{R}^{6}$ is the external force in Cartesian space.

A desired impedance model \cite{bruno2010roboticsBookSiciliano, song2019tutorialSurveyImpedanceControl,teng4IROS2022NDOB} for robot-environment interaction can be expressed as
\begin{equation}
\label{eqn_ImpedModelFull_augedwithcc}
\begin{aligned}
   \mathbf{F_{imp}} 
   &= \mathbf{M_{imp}(\ddot{x}-\ddot{x}_d)} \\
   &\quad\quad \mathbf {+ (S_x + D_{imp}) (\dot{x}-\dot{x}_d) + K_{imp} (x-x_{d})
    }\\
\end{aligned}
\end{equation}
\noindent
where $\mathbf{M_{imp}, D_{imp}, K_{imp}}$ are user-designed matrices for inertia, damping, and stiffness, respectively. Note that $\mathbf{x_d, \dot{x}_d, \ddot{x}_d}$ are the desired position, velocity, and acceleration, respectively in Cartesian space, while $\mathbf{x, \dot{x}, \ddot{x}}$ are the actual ones. $\mathbf{S_x}$ is the Coriolis and centrifugal matrix of the robot in Cartesian space and $\mathbf{S_x = J^{-T} S J^{-1} - M_x \dot{J} J^{-1}}$, where $\mathbf{M_x = J^{-T} M J^{-1}}$ is the inherent inertia of the robot in Cartesian space \cite{12torabi2019applicationAliIRAL}.

To avoid the measurement of external forces \cite{teng4IROS2022NDOB}, the designed inertia matrix can be set as the inherent inertia matrix of the robot, \ie{}, $\mathbf{M_{imp} = M_x}$. Then, to reach \eqref{eqn_ImpedModelFull_augedwithcc} as the closed-loop dynamics governing the robot-environment interaction in an ideal scenario of no model errors and no joint friction, setting with $\mathbf{F_{imp} = F_{ext}}$, the impedance control law can be given by 
\begin{equation}
\label{eqn_imped_controller_simplify2}
\begin{aligned}
    \mathbf{\boldsymbol{\tau}} &=
    \mathbf{
    M J^{-1} (\ddot{x}_d - \dot{J} J^{-1} \dot{x}_d ) + S J^{-1} \dot{x}_d + g }\\
    &\quad\quad \mathbf{+ J^T [D_{imp} (\dot{x}_d - \dot{x}) + K_{imp} (x_d-x)] } \\ 
\end{aligned}
\end{equation}

Note that when implementing the impedance controller (\ref{eqn_imped_controller_simplify2}) in practice, the estimates $\mathbf{\hat{M}, \hat{S}, \hat{g}}$ will be used for the calculation since an accurate model of a physical robot is usually not available.

For robot end-effector (EE) moving to a fix point, \ie{}, set-point regulation, it has $\mathbf{\ddot{x}_d=0}$, $\mathbf{\dot{x}_d=0}$. Then, the impedance control law (\ref{eqn_imped_controller_simplify2}) can be simplified to \eqref{eqn_imped_controller_simplify3}, which is also known as task-space proportional–derivative (PD) controller with gravity compensation. 
\begin{equation}
\label{eqn_imped_controller_simplify3}
  \mathbf{
  \boldsymbol{\tau} = J^T [K_{imp} (x_{d}-x) - D_{imp} \dot{x}] + g }
\end{equation}

Based on (\ref{eqn_imped_controller_simplify3}), when the fixed point is set to be time-varying points of the real-time Cartesian position, \ie{}, $\mathbf{x_d=x}$, it can be further reduced to be 
\begin{equation}
\label{eqn_imped_controller_simplify4_pHRI}
  \mathbf{
  \boldsymbol{\tau} = - J^T D_{imp} \dot{x} + g}
\end{equation}
\noindent
in which mode, physical human-robot interaction ($p$HRI) is enabled, and the user can move the robot EE freely in its workspace. Note that in this $p$HRI mode, it is possible to involve force feedback if needed, then the controller with force feedback will be given by
\begin{equation}
\label{eqn_imped_controller_simplify4_pHRI_ff}
  \mathbf{
  \boldsymbol{\tau} = - J^T D_{imp} \dot{x} + g + \boldsymbol{\tau} \mathbf{_{ff}}}
\end{equation}
\noindent
where the $\boldsymbol{\tau} \mathbf{_{ff}} $ is the rendered force feedback which will be introduced in detail later. It is worth mentioning that for a backdrivable haptic device, it is defaulted in $p$HRI mode without needing a controller like (\ref{eqn_imped_controller_simplify4_pHRI}), and the operator feels the intrinsic inertia of the robot mechanism. In other words, only the force feedback term $\boldsymbol{\tau} \mathbf{_{ff}} $ in (\ref{eqn_imped_controller_simplify4_pHRI_ff}) is needed for a haptic device if force feedback rendering is needed, while the damper term is applicable but not necessary. If the operator implements the whole controller (\ref{eqn_imped_controller_simplify4_pHRI_ff}) on a haptic device for $p$HRI with force feedback rendering, the operator will feel an extra damper force (due to the damper term) in addition to the intrinsic inertia from the robot mechanism, but the damper force is usually small since robot EE held by the operator is usually in a low-speed movement. The advantage of the controller (\ref{eqn_imped_controller_simplify4_pHRI_ff}) is that it is a general form for robotic control thus applicable to all robotic systems other than haptic devices. Furthermore, it can realize a seamless transition from trajectory-tracking tasks to $p$HRI due to the fact that (\ref{eqn_imped_controller_simplify4_pHRI}) ($p$HRI mode) is automatically reduced from (\ref{eqn_imped_controller_simplify2}) (trajectory-tracking mode) under some $p$HRI-specific conditions. 

\begin{figure*}[!t] 
    \centering
    \setlength{\fboxsep}{5pt}
    \fcolorbox{white}{white}{
    \includegraphics[width=0.98\textwidth]{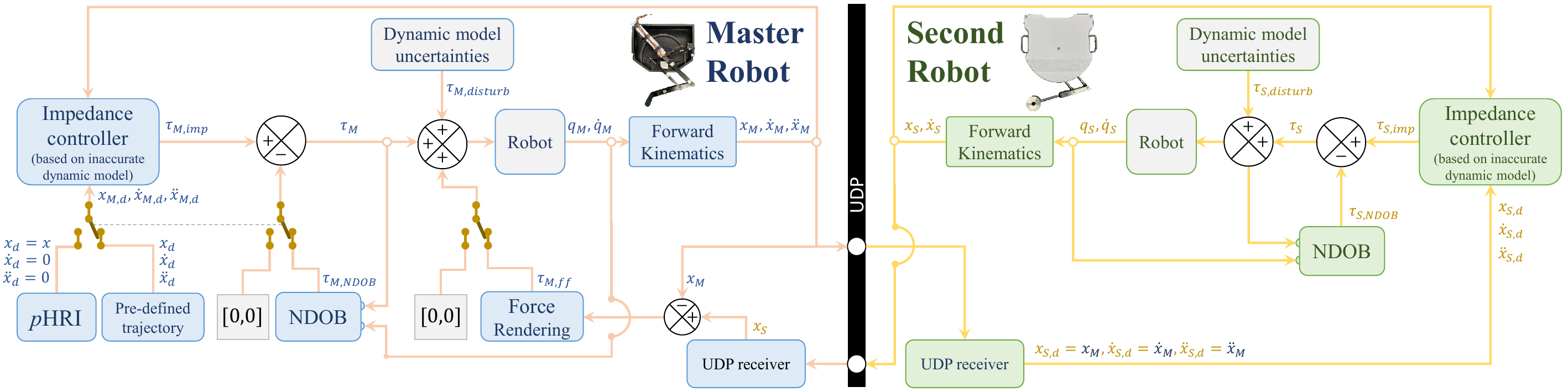}
    }
    \caption{Block diagram of control schemes for the proposed teleoperation system. The dashed line means a linked switch. The output of NDOB $\tau_{\tt M,NDOB}$ and $\tau_{\tt S,NDOB}$ are the estimation of the lumped uncertainties $\tau_{\tt M,disturb}$ and $\tau_{\tt S,disturb}$, respectively. } 
    \label{fig02_control_block_diagram_NDOB}
\end{figure*}

For the teleoperation system proposed in this paper, as illustrated in Fig.\ref{fig02_control_block_diagram_NDOB}, both the master robot and the second robot are implemented with an impedance controller based on (\ref{eqn_imped_controller_simplify2}) which are expressed as
\begin{equation}
\label{eqn_imped_controller_simplify2_MS}
\begin{aligned}
\begin{cases}
    \mathbf{\boldsymbol{{\tau}}_{M,imp}} \\
    =
    \mathbf{
    \hat{M} J^{-1} (\ddot{x}_{M,d} - \dot{J} J^{-1} \dot{x}_{M,d} ) + \hat{S} J^{-1} \dot{x}_{M,d} + \hat{g}_M +\boldsymbol{\tau} \mathbf{_{ff}} }\\
    \quad\quad \mathbf{+ J^T [D_{M,imp} (\dot{x}_{M,d} - \dot{x}_M) + K_{M,imp} (x_{M,d}-x_M)] }\\
    \mathbf{\boldsymbol{{\tau}}_{S,imp}} \\
    =
    \mathbf{
    \hat{M} J^{-1} (\ddot{x}_{S,d} - \dot{J} J^{-1} \dot{x}_{S,d} ) + \hat{S} J^{-1} \dot{x}_{S,d} + \hat{g}_S }\\
    \quad\quad \mathbf{+ J^T [D_{S,imp} (\dot{x}_{S,d} - \dot{x}_S) + K_{S,imp} (x_{S,d}-x_S)] }\\
\end{cases}
\end{aligned}
\end{equation}
\noindent
where the subscript $\mathbf{_M}$ and $\mathbf{_S}$ stand for ``Master'' and ``Second'', respectively, and subscript $\mathbf{_{imp}}$ stands for ``impedance'', and subscript $\mathbf{_{d}}$ stands for ``desired''. It should be noted that the force feedback $\boldsymbol{\tau} \mathbf{_{ff}}$ is only set for the master robot in $p$HRI mode.

\subsection{Force Feedback and Rendering}

The proposed teleoperation system, as shown in Fig.\ref{fig02_control_block_diagram_NDOB}, is set as bilateral, which means that force feedback is rendered and delivered to the operator on the master robot side when the second robot interacts with the surrounding environment. In this work, a virtual spring force will be rendered as the force feedback based on position error between the two robots. To this end, the real-time position of the second robot is sent back to the master robot via user datagram protocol (UDP) for force rendering. The position error between the master robot EE and the second robot EE is calculated for generating a virtual spring force which can be expressed as
\begin{equation}
\label{eqn_Fkx_force_feedback_rendering}
\begin{aligned}
    \boldsymbol{\tau}\mathbf{_{ff}} =\mathbf{J^T F_{ff}} = \mathbf{J^T K_{ff} (x_S - x_M)}
\end{aligned}
\end{equation}
\noindent
where $(\mathbf{x_S - x_M})$ is the position error between the master robot EE and the second robot EE, and $\mathbf{K_{ff}}$ is the virtual spring stiffness coefficient which can be tuned to make the virtual spring stiffer or softer.

\subsection{Disturbance and Disturbance Observer}

By collecting all the disturbances together, the dynamic model (\ref{eqn_dynModel_general}) of a robot without considering force feedback ($\boldsymbol{\tau} \mathbf{_{ff}=0}$) can be re-written as
\begin{equation}
 \label{eqn_dynModel_general2}
 \begin{aligned}
    \mathbf{ \hat{M} \ddot{q} +\hat{S} \dot{q} +\hat{g}} 
    = \mathbf{\boldsymbol{\tau} + }
    \underbrace{\mathbf{\boldsymbol{\tau}_{ext}- [\boldsymbol{\tau}_{fric}+(\varDelta M \ddot{q}+\varDelta S \dot{q}+\varDelta g)]}}_\mathbf{\boldsymbol{\tau}_{disturb}}
\end{aligned}
\end{equation}
\noindent
where $\mathbf{\boldsymbol{\tau}_{disturb}}$ denotes the lumped uncertainties that usually include three main aspects, \ie{}, the model error $\mathbf{(\varDelta M \ddot{q}+\varDelta S \dot{q}+\varDelta g)}$, the joint friction $\mathbf{\boldsymbol{\tau}_{fric}}$, and the external disturbances $\mathbf{\boldsymbol{\tau}_{ext}}$, where the last aspect may involve constant disturbance and/or time-varying disturbance. The constant disturbance may be a constant payload attached to the robot end-effector (EE) or body, while time-varying disturbance may be robot-environment interaction forces such as human-applied forces during human-robot interaction. A disturbance observer usually estimates the lumped uncertainties $\mathbf{\boldsymbol{\tau}_{disturb}}$ \cite{mohammadi2013nonlinearNDOBtavakoli}, but cannot discriminate any single component when more than one component exists.

The main uncertainties of the teleoperation system (in free motion mode) constructed in this paper will include the model error and the inaccurate joint friction which will be estimated and compensated for by using a disturbance observer. A variety of observers are available for use \cite{radke2006surveyAllOBS, chen2016disturbanceOverviewSurvey,mohammadi2017nonlinearNDOBreview,haddadin2017robotDOBsurveyDeLuca}, such as generalized momentum observer (GMO) \cite{haddadin2017robotDOBsurveyDeLuca}, extended state observer (ESO) \cite{sebastian2019RALinteractionESOimproved}, nonlinear disturbance observer (NDOB) \cite{mohammadi2013nonlinearNDOBtavakoli}, and disturbance Kalman filter (DKF) method \cite{huTIE2018contactDKF, liu2021KTHsensorlessDKF}.

In this paper, a nonlinear disturbance observer (NDOB) is employed considering that the NDOB has the advantage of estimating the nonlinearities in the dynamics. An adapted NDOB design based on \cite{mohammadi2013nonlinearNDOBtavakoli} can be expressed as
\begin{equation}
 \label{eqn_NDOB_observer}
 \begin{cases}
    \mathbf{L_{obs} = Y_{obs} {M}_{obs}^{-1}} \\
    \mathbf{p = Y_{obs} \dot{q}} \\
    \mathbf{\dot{z} = -L_{obs} z + L_{obs} (\hat{S} \dot{q} + \hat{g} - \tau - p)} \\
    \mathbf{\boldsymbol{\tau}_{\tt {NDOB}} = z + p} \\
 \end{cases}
\end{equation}
\noindent
where $\mathbf{L_{obs}} \in \mathbb{R}^{n \times n}$ is the observer gain matrix, $\mathbf{Y_{obs}} \in \mathbb{R}^{n \times n}$ is a constant invertible matrix needs to be designed, $\mathbf{{M}_{obs}}$ is designed to be a symmetric and positive definite matrix in practice and thus invertible, $\mathbf{z}$ is an auxiliary variable, $\mathbf{p}$ is an auxiliary vector determined from $\mathbf{Y_{obs}}$, $\mathbf{\boldsymbol{\tau}_{\tt {NDOB}}}$ is the estimated lumped uncertainties via the NDOB observer. For the full designing procedures of this NDOB and complete theoretical analysis, refer to \cite{mohammadi2013nonlinearNDOBtavakoli}.

A teleoperation system integrating impedance control with NDOB is proposed for robot-assisted rehabilitation. The block diagram of the proposed teleoperation system is illustrated in Fig.\ref{fig02_control_block_diagram_NDOB}.

\section{Robotic Systems for Teleoperation}
\label{robots}

\begin{figure}
    \centering
    \setlength{\fboxsep}{5pt} 
    \fcolorbox{white}{white}{
    \includegraphics[width=0.4\textwidth]{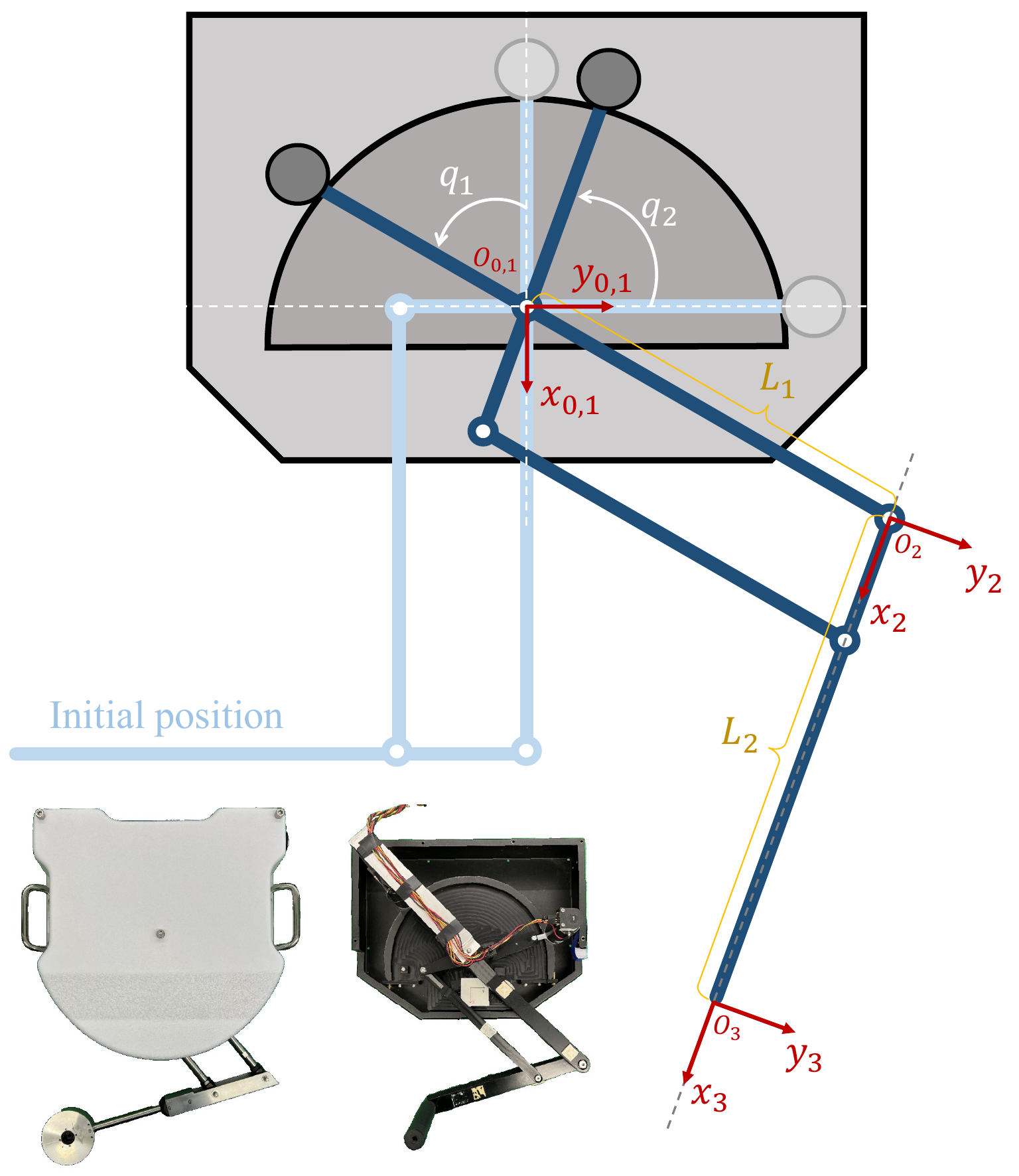}
    }
    \caption{Schematic of the 2-DOF planar upper-limb rehabilitation robot (black, white) and frame attachment to each joint. Frame \{0\} is the base frame while frame \{3\} is the end-effector (EE) frame. $L_1, L_2$ are link lengths. $q_1, q_2$ are joint angle variables.} 
    \label{fig1_2dofRehab_schematic_frames}
\end{figure}

The experimental setup is shown in Fig.~\ref{fig1_2dofRehab_schematic_frames}. A 2-DOF planar upper-limb rehabilitation robot 1.0 (in black color, Quanser Inc., Markham, ON, Canada), and a 2-DOF planar upper-limb rehabilitation robot 2.0 (in white color, Quanser Inc., Markham, ON, Canada) are used for constructing the teleoperation system. The Rehab robot 1.0 (black) is used as the master robot while the Rehab robot 2.0 (white) is used as the second robot. To build virtual models of the two robots, the kinematic model and dynamic model of the two robots are reconstructed based on \cite{ccavucsouglu2002criticalPhantomPremium, dyck2013measuringMScThesis, torabi2020usingTeleBlackWhite2DOF}. It should be noted that the Rehab robot 2.0 (white) is an upgraded version of the Rehab robot 1.0 (black), so they have the same form of kinematics and dynamics (with different values for the parameters). All the experiments are conducted using MATLAB/Simulink (version R2017a, MathWorks Inc., Natick, MA, USA) with Quarc real-time control software (Quanser Inc., Markham, ON, Canada), which is running on two computers with a 3.33 GHz Intel(R) Core(TM) 2 i5 CPU and a Windows 7 Enterprise 64-bit operating system. The control rate of the robots is $1,000$ Hz, while the sampling rate for acquiring data is set as $500$ Hz. The bilateral communications between the two robots are realized by UDP at a rate of $1,000$ Hz.

\subsection{Robot Kinematics}

A thorough exploration of the kinematics and dynamics of the 2-DOF rehabilitation robot 1.0 (black) can be found in \cite{dyck2013measuringMScThesis, torabi2020usingTeleBlackWhite2DOF}. Additionally, a Denavit–Hartenberg (DH) table for this robot is provided in Table \ref{tab1_DH_2dofRehab}. The corresponding frame definitions are illustrated in Fig. \ref{fig1_2dofRehab_schematic_frames}. Note that the DH table and frame definitions are derived based on the Rehab robot 1.0 (black), but they are also applied to the Rehab robot 2.0 (white).

\begin{table} 
    \caption{Denavit–Hartenberg (DH) parameters for the {2-DOF} Quanser Rehab robot 1.0 (black) and Rehab robot 2.0 (white) kinematic chain (for the homogeneous transform in the modified convention).} 
    \centering 
    \scalebox{1.0}{
    \begin{threeparttable}[t]
    \begin{tabular}{ cc l c l c l}
    \toprule
    & no. & Joint & $a(m)$ & $\alpha(rad)$ & $d(m)$ & $\theta(rad)$  \\ 
    \midrule 
    & 1 & Joint 1       & 0      & 0  & 0  & $q_1$  \\
    & 2 & Joint 2       & $L_1$  & 0  & 0  & $-\frac{\pi}{2} - q_1 + q_2$  \\
    & 3 & Joint 3 (EE)  & $L_2$  & 0  & 0  & 0  \\
    \bottomrule 
    \end{tabular}
    \end{threeparttable}
    }
    \label{tab1_DH_2dofRehab} 
\end{table}

According to the DH parameters in Table \ref{tab1_DH_2dofRehab} and the frames determined in Fig. \ref{fig1_2dofRehab_schematic_frames}, the homogeneous transformation matrix $\mathbf{T}$ from EE frame \{3\} to base frame \{0\} can be obtained as
\begin{equation}
\label{eqn_T0ee}
\begin{aligned}
    \mathbf{T}
    &= \left[\begin{array}{rrrr} 
        \sin(q_2),& \cos(q_2),& 0,& L_1 \cos(q_1) + L_2 \sin(q_2) \\
        -\cos(q_2),& \sin(q_2),& 0,& L_1 \sin(q_1) - L_2 \cos(q_2) \\
        0 ,&  0 ,&  1 ,&  0 \\
        0 ,&  0 ,&  0 ,&  1 \\
    \end{array}\right]
\end{aligned}
\end{equation}
\noindent
where $q_1, q_2$ are joint angles, $L_1, L_2$ are link lengths.

The Jacobian matrix $\mathbf{J}$ can be expressed \cite{dyck2013measuringMScThesis, torabi2020usingTeleBlackWhite2DOF} by
\begin{equation}
\label{eqn_Jacobian_spatial_2DOF}
\begin{aligned}
    \mathbf{J}
    &= \left[\begin{array}{rr} 
        -L_1 \sin(q_1),& L_2 \cos(q_2) \\
        L_1 \cos(q_1),& L_2 \sin(q_2) \\
    \end{array}\right]
\end{aligned}
\end{equation}
\noindent
where $q_1, q_2$ are joint angles, $L_1, L_2$ are link lengths.

Based on the transformation matrix (\ref{eqn_T0ee}), the forward kinematics \cite{dyck2013measuringMScThesis, torabi2020usingTeleBlackWhite2DOF} can be written as
\begin{equation}
 \label{eqn_kinematics_forward}
 \begin{cases}
     x &= L_1 \cos(q_1) + L_2 \sin(q_2) \\
     y &= L_1 \sin(q_1) - L_2 \cos(q_2) \\
 \end{cases}
\end{equation}

For the Rehab robot 1.0 (black), the link lengths are $L_1 = 0.254$ m, $L_2 =0.2667$ m. The joint limits restricted by the mechanical structure \cite{dyck2013measuringMScThesis} are
\begin{equation}
 \label{eqn_joint_limits_2DOF_black}
 \begin{cases}
     -55\degree \leq q_1 \leq 90\degree \\
     0\degree \leq q_2 \leq 145\degree \\
     35\degree \leq q_1 - q_2 + 90\degree \leq 145 \degree \\
 \end{cases}
\end{equation}

For the Rehab robot 2.0 (white), the link lengths are $L_1 = 0.340$ m, $L_2=0.375$ m. The joint limits restricted by the mechanical structure are
\begin{equation}
 \label{eqn_joint_limits_2DOF_white}
 \begin{cases}
     -86\degree \leq q_1 \leq 132\degree \\
     -49\degree \leq q_2 \leq 154\degree \\
     35\degree \leq q_1 - q_2 + 90\degree \leq 145 \degree \\
 \end{cases}
\end{equation}

The inverse kinematics is obtained \cite{dyck2013measuringMScThesis} as
\begin{equation}
 \label{eqn_kinematics_inverse}
 \begin{cases}
     q_1 = acos(\frac{x^2+y^2+L^2_1-L^2_2}{2 L_1 \sqrt{x^2+y^2}}) + atan2(y,x) \\
     q_2 = q_1 + acos(\frac{L^2_1+L^2_2-x^2-y^2}{2 L_1 L_2}) - 90\degree \\
 \end{cases}
\end{equation}

\begin{figure}
    \centering
    \includegraphics[width=0.48 \textwidth]{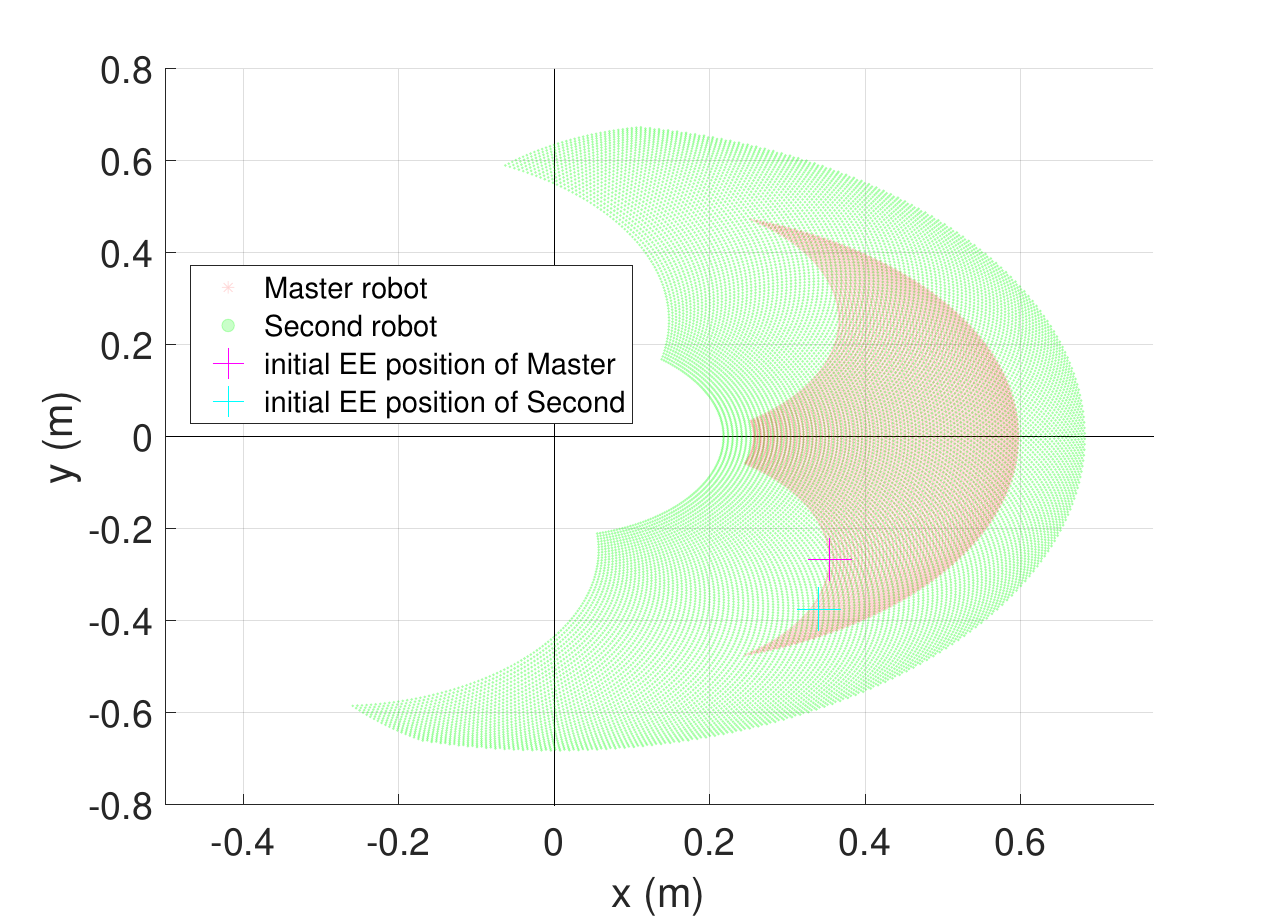}
    \caption{Workspace of the master robot and the second robot.} 
    \label{fig01_workspace_2DOFblack_2DOFwhite}
\end{figure}

The workspace of the master robot and the second robot are illustrated in Fig.~\ref{fig01_workspace_2DOFblack_2DOFwhite}. Note that the workspace of the master robot (the red color area) is shifted $+0.1$ m along the $x$-axis for a better overlap with the workspace of the second robot.

\subsection{Robot Identified-Dynamics}

According to \cite{dyck2013measuringMScThesis}, the identified inertia matrix for the Rehab robot 1.0 (black) is given by
\begin{equation}
\label{eqn_M_inertia}
\begin{aligned}
    \mathbf{\hat{M}}
    &= \left[\begin{array}{rr} 
        \alpha_1,& -\frac{1}{2} \alpha_2 \sin(q_1-q_2) \\
        -\frac{1}{2} \alpha_2 \sin(q_1-q_2),& \alpha_3 \\
    \end{array}\right]
\end{aligned}
\end{equation}
\noindent
where $\alpha_1$=0.06929, $\alpha_2$=0.04217, $\alpha_3$=0.04416 are experimentally identified dynamic coefficients in the inertia matrix.

The identified Coriolis and centrifugal forces related matrix for the Rehab robot 1.0 (black) \cite{dyck2013measuringMScThesis} is given by
\begin{equation}
\label{eqn_S_coriolis_centrifugal}
\begin{aligned}
    \mathbf{\hat{S}}
    &= \left[\begin{array}{rr} 
        0,& \frac{1}{2} \alpha_2 \cos(q_1-q_2) \dot{q}_2 \\
        \frac{1}{2} \alpha_2 \cos(q_1-q_2) \dot{q_1},& 0 \\
    \end{array}\right]
\end{aligned}
\end{equation}
\noindent
where $\alpha_2$=0.04217 is experimentally identified.

As the Rehab robot 1.0 (black) is a planar robot and its links move in the horizontal plane, the gravity-related vector \cite{dyck2013measuringMScThesis} is simply given by
\begin{equation*}
\begin{aligned}
    \mathbf{\hat{G}}
    &= \left[\begin{array}{r} 
        0 \\
        0 \\
    \end{array}\right]
\end{aligned}
\end{equation*}

The identified joint friction model for the Rehab robot 1.0 (black) \cite{dyck2013measuringMScThesis} is given by
\begin{equation}
\label{eqn_friction}
\begin{aligned}
    \boldsymbol{\hat{\tau}}\mathbf{_{fric}}
    &= \left[\begin{array}{r} 
        \alpha_4 \dot{q}_1 \\
        \alpha_5 \dot{q}_2 \\
    \end{array}\right]
\end{aligned}
\end{equation}
\noindent
where $\alpha_4$=0.06510, $\alpha_5$=0.07389 are experimentally identified.

For the Rehab robot 2.0 (white), the same set of the identified dynamics model presented above is employed. Although the actual identified dynamic parameters of the Rehab robot 2.0 (white) would be quite different from the black one, it is possible to use the same dynamic model since a disturbance observer is employed to compensate for the dynamic uncertainties.

\subsection{Parameterization}

\begin{table}
    \caption{Parameterization for experiments with the teleoperation system for robot-assisted rehabilitation.} 
    \centering 
    \begin{tabular}{ llll } 
    \toprule
    Parameters & Master Robot & Second Robot & Location \\ 
    \midrule 
    Link length  & $[0.254,0.2667]$ & $[0.340,0.375]$ & Eqn.(\ref{eqn_T0ee},\ref{eqn_Jacobian_spatial_2DOF},\ref{eqn_kinematics_forward},\ref{eqn_kinematics_inverse}) \\
    Spring  & $\mathbf{K_{imp} = 30 I}$ & $\mathbf{K_{imp} = 20 I}$ & Eqn.(\ref{eqn_imped_controller_simplify2_MS}) \\
    Damping  & $\mathbf{D_{imp} = 2 \sqrt{30} I}$ & $\mathbf{D_{imp} = 2 \sqrt{20} I}$ & Eqn.(\ref{eqn_imped_controller_simplify2_MS}) \\
    Inertia matrix & $\mathbf{{M}_{obs} = \hat{M}}$ & $\mathbf{{M}_{obs} = 0.001 I}$ & Eqn.(\ref{eqn_NDOB_observer}) \\
    Observer gain & $\mathbf{Y_{obs} = 1.92 I}$ & $\mathbf{Y_{obs} =0.048 I}$ & Eqn.(\ref{eqn_NDOB_observer}) \\
    \bottomrule 
    \end{tabular}
        \begin{tablenotes}
        \small
        \item Note: $\mathbf{I} \in \mathbb{R}^{2 \times 2} $ denote identity matrix. 
        \end{tablenotes}
    \label{tab2_parameterization_exp} 
\end{table}

For all experiments in the remaining part of this paper, the parameter values used in the impedance model and NDOB are listed in Table \ref{tab2_parameterization_exp}. Note that the parameter values are tuned by trial and error with the strategy of binary search. A video\footnote{Online video link: \url{https://youtu.be/Jokv_RPOXEc}} demonstration of the experiments is available \href{https://youtu.be/Jokv_RPOXEc}{online}.

In robot-assisted rehabilitation, a trajectory, either pre-defined by mathematical equations or customized by the therapist, is needed to help patients move their injured limbs around for rehabilitation and recovery \cite{sharifi2018patientJMR, sharifi2020assistAsNeededCEP, najafi2020usingPotentialFieldTMECH}. Several pre-defined trajectories are employed for evaluating the teleoperation system performance while trajectory tracking accuracy will be used to evaluate the system performance. Note that in order to present the results more intuitively, the tracking accuracy in experimental results is presented as trajectory plots by overlapping the desired and the actual trajectories into one figure instead of trajectory tracking error plots. A circular and cyclic trajectory can be expressed as a function of time as the following
\begin{equation}
\label{eqn_trajectory_0}
\begin{cases}
    x_{d} = R \sin(\frac{2\pi}{t_1} t) \\
    y_{d} = R \cos(\frac{2\pi}{t_1} t) \\
\end{cases}
\end{equation}
\noindent
where $R=0.08$ m is the radius of the circle and $t_1 = 5$ s is the period for generating a full cycle.

A figure-eight trajectory can be expressed as 
\begin{equation}
\label{eqn_trajectory_8}
\begin{cases}
    x_{d} = R \sin(\frac{2\pi}{t_1} t) \cos(\frac{2\pi}{t_1} t) \\
    y_{d} = R \sin(\frac{2\pi}{t_1} t) \\
\end{cases}
\end{equation}
\noindent
where $R=0.1$ m is the amplitude of the figure-eight trajectory, $t_1 = 5$ s is the period for generating a full cycle.

A tetragon trajectory (also known as beetle curve) can be expressed as a function of time given by 
\begin{equation}
\label{eqn_trajectory_4}
\begin{cases}
    x_{d} = R \cos^3(t) \\
    y_{d} = R \sin^3(t) \\ 
\end{cases}
\end{equation}
\noindent
where $R=0.1$ m is a parameter of the tetragon trajectory. 

A pentagram trajectory (also known as hypotrochoid) can be expressed as a function of time given by 
\begin{equation}
\label{eqn_trajectory_5}
\begin{cases}
    x_{d} = (R-r) \cos(n t) + d \cos(\frac{R-r}{r} n t) \\
    y_{d} = (R-r) \sin(n t) - d \sin(\frac{R-r}{r} n t) \\
\end{cases}
\end{equation}
\noindent
where $R=0.08$, $r=0.048$, $d=0.064$, and $n=3$, are parameters of the pentagram trajectory, and $R$, $r$, $d$ control the size of the trajectory, and $n$ controls the period time for a full cycle.

A rose curve trajectory (also known as rhodonea curve) can be expressed as a function of time given by 
\begin{equation}
\label{eqn_trajectory_8888}
\begin{cases}
    x_{d} = R \cos(k t) \cos(t) \\
    y_{d} = R \cos(k t) \sin(t) \\
\end{cases}
\end{equation}
\noindent
where $R=0.1$, $k=4$, are parameters of the rose curve trajectory, and $R$ controls the size of pattern, and $k$ controls the number of petals of the pattern.

\section{Experiments and Results}
\label{experiments}

\subsection{Exp.1: Individual Robot Performance}

In Exp.1, we evaluated the performance of each robot by implementing the impedance controller (\ref{eqn_imped_controller_simplify2}) with and without the nonlinear disturbance observer (NDOB) (\ref{eqn_NDOB_observer}). In this experiment, a circular trajectory (\ref{eqn_trajectory_0}) is employed.

\begin{figure}
     \centering 
     \begin{subfigure}[b]{0.24\textwidth}
         \centering
         \includegraphics[width=\textwidth]{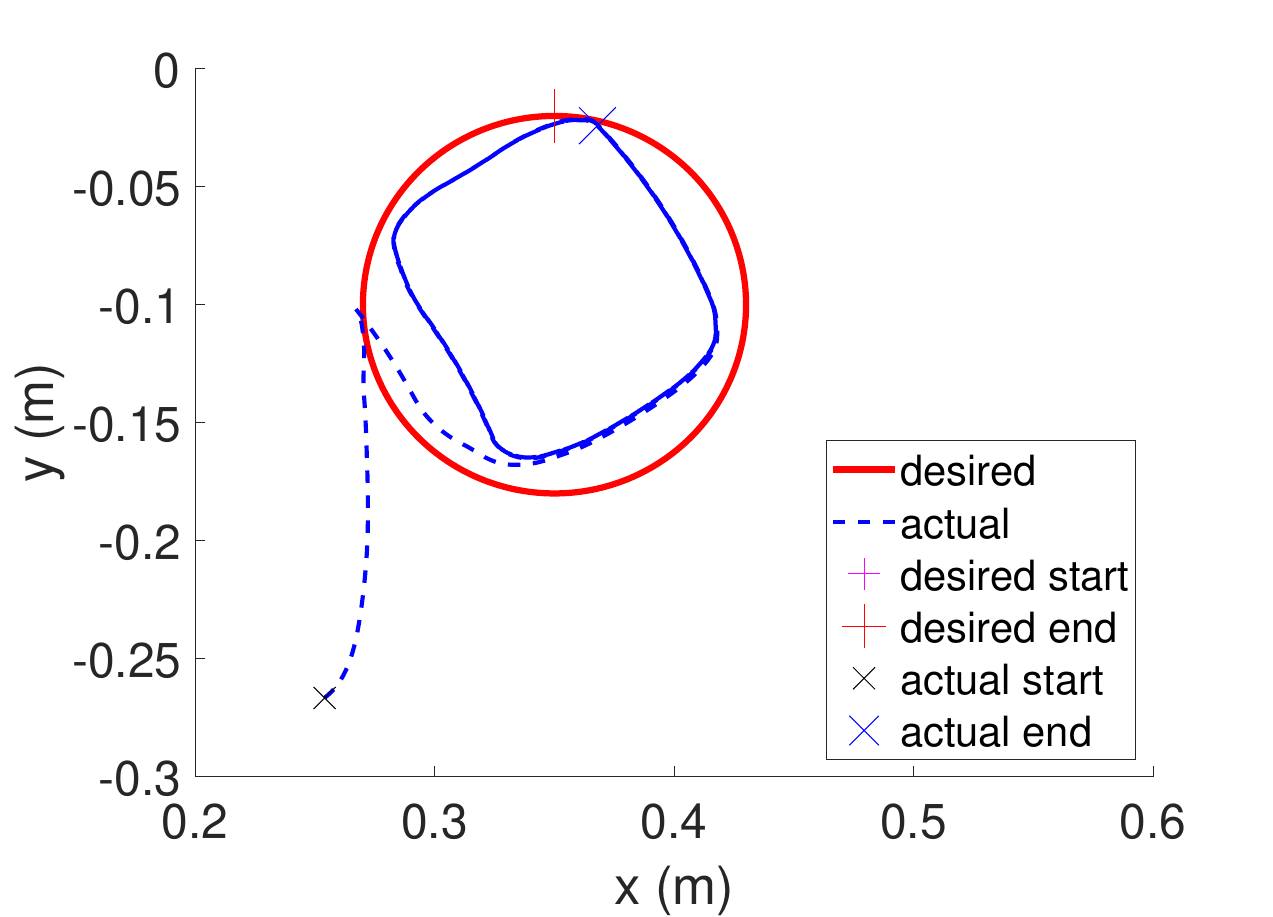}
         \caption{Master robot without NDOB}
         \label{fig04a_trajectory_exp00_ref}
     \end{subfigure}
     \hfill
     \begin{subfigure}[b]{0.24\textwidth}
         \centering
         \includegraphics[width=\textwidth]{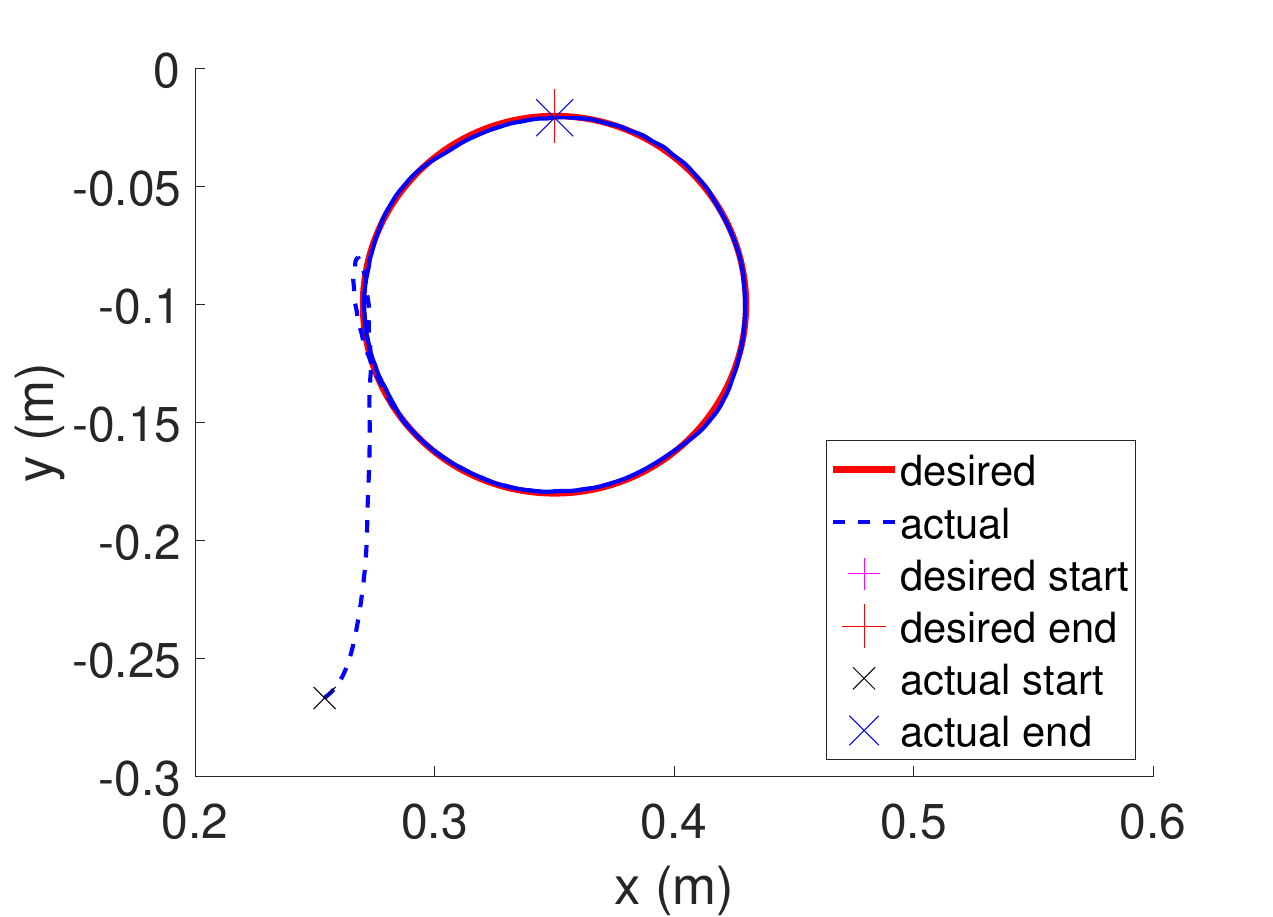}
         \caption{Master robot with NDOB}
         \label{fig04b_trajectory_exp00_ref}
     \end{subfigure}
     \hfill
     \begin{subfigure}[b]{0.24\textwidth}
         \centering
         \includegraphics[width=\textwidth]{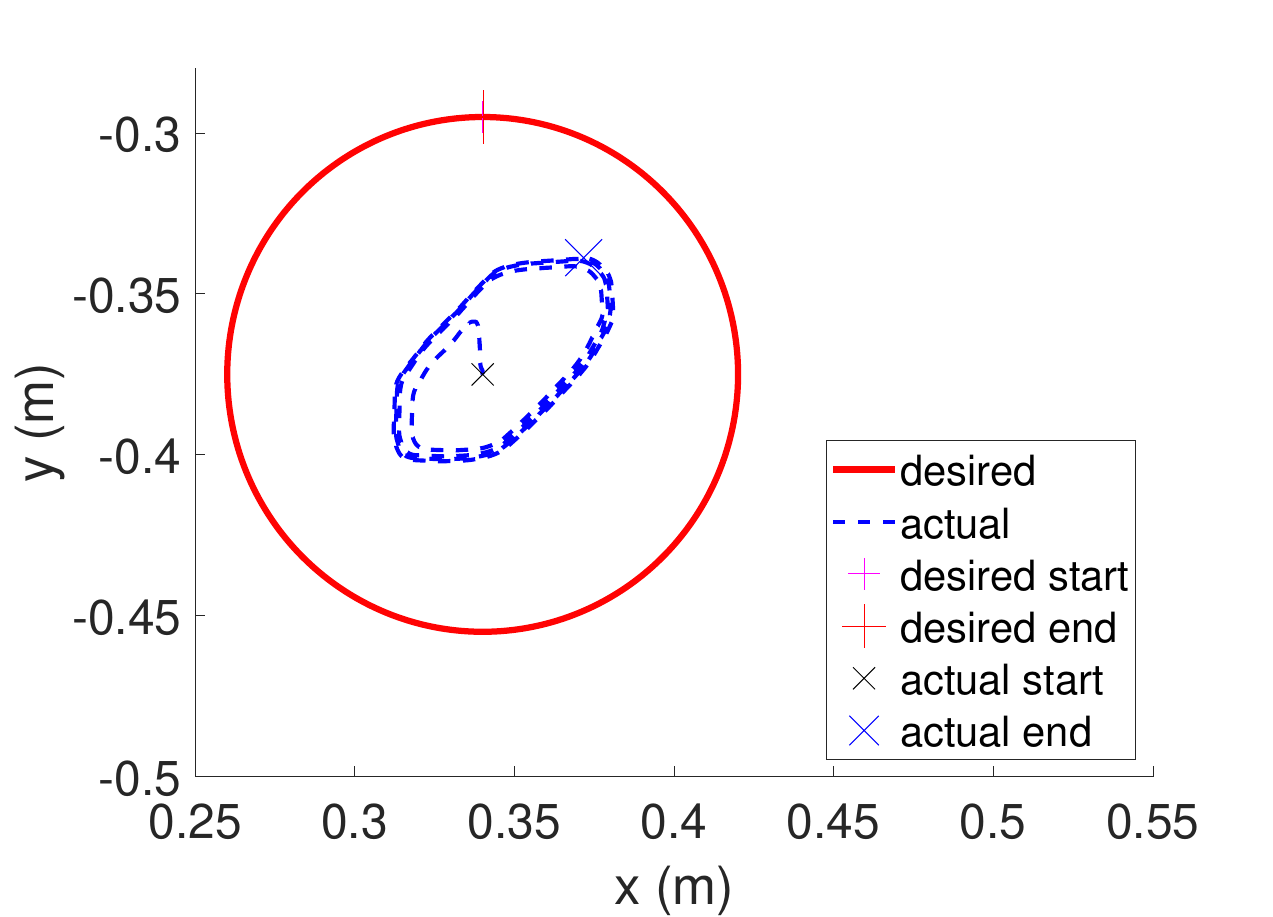}
         \caption{Second robot without NDOB}
         \label{fig04c_trajectory_exp00_ref}
     \end{subfigure}
     \hfill
     \begin{subfigure}[b]{0.24\textwidth}
         \centering
         \includegraphics[width=\textwidth]{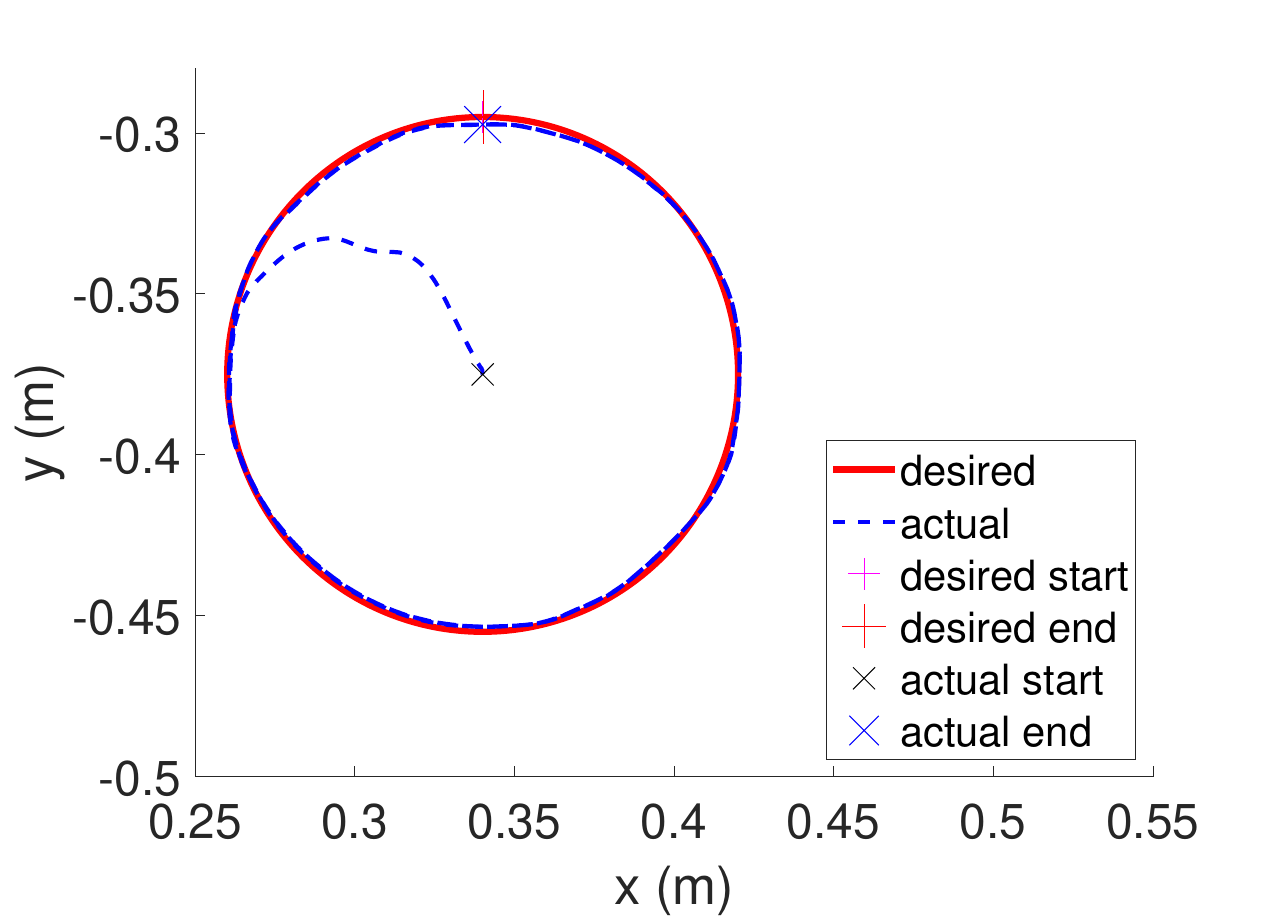}
         \caption{Second robot with NDOB}
         \label{fig04d_trajectory_exp00_ref}
     \end{subfigure}
        \caption{Exp.1 results of individual performance of the two robots by implementing impedance controller with or without NDOB. }
        \label{fig04_trajectory_exp00_ref}
\end{figure}

The experiment results are shown in Fig.~\ref{fig04_trajectory_exp00_ref}. In the figure, we can see that the robots cannot perform the trajectory tracking task accurately without NDOB (Fig.~\ref{fig04a_trajectory_exp00_ref},~\ref{fig04c_trajectory_exp00_ref}), while accurate tracking performance is obtained (Fig.~\ref{fig04b_trajectory_exp00_ref},~\ref{fig04d_trajectory_exp00_ref}) when NDOB is implemented. The reason for failed-tracking (in Fig.~\ref{fig04a_trajectory_exp00_ref},~\ref{fig04c_trajectory_exp00_ref}) is due to the inaccurate dynamic model and joint friction model of the robot. This is especially true for the second robot (white) since the identified dynamic model and friction model were specially for the master robot (black). With implementing an NDOB observer, the dynamic uncertainties can be accurately estimated and compensated for in both robots.

Exp.1 indicates that by implementing both the impedance controller and NDOB, the robots can accurately perform trajectory-tracking tasks separately. Even with a borrowed dynamic model, which can be viewed as a roughly estimated dynamic model for the second robot (white), the robot can track accurately by involving a disturbance observer.

\subsection{Exp.2: Trajectory Tracking}

In Exp.2, a set of pre-defined trajectories with patterns varying from simple to complex are employed to evaluate the tracking performance of the teleoperation system, including a circle (\ref{eqn_trajectory_0}), a figure-eight trajectory (\ref{eqn_trajectory_8}), a tetragon trajectory (\ref{eqn_trajectory_4}), a pentagram trajectory (\ref{eqn_trajectory_5}), and a rose curve trajectory (\ref{eqn_trajectory_8888}). Both the master and the second robots are implemented with an impedance controller and NDOB observer. The master robot is set to track the pre-defined trajectories, while the second robot is set to track the EE position of the master robot. The teleoperation control in Exp.2 is unilateral, \ie{}, no force feedback on the master robot side, since force feedback is not necessary for the master robot in this experiment.

\begin{figure}
     \centering
     \begin{subfigure}[b]{0.24\textwidth}
         \centering
         \includegraphics[width=\textwidth]{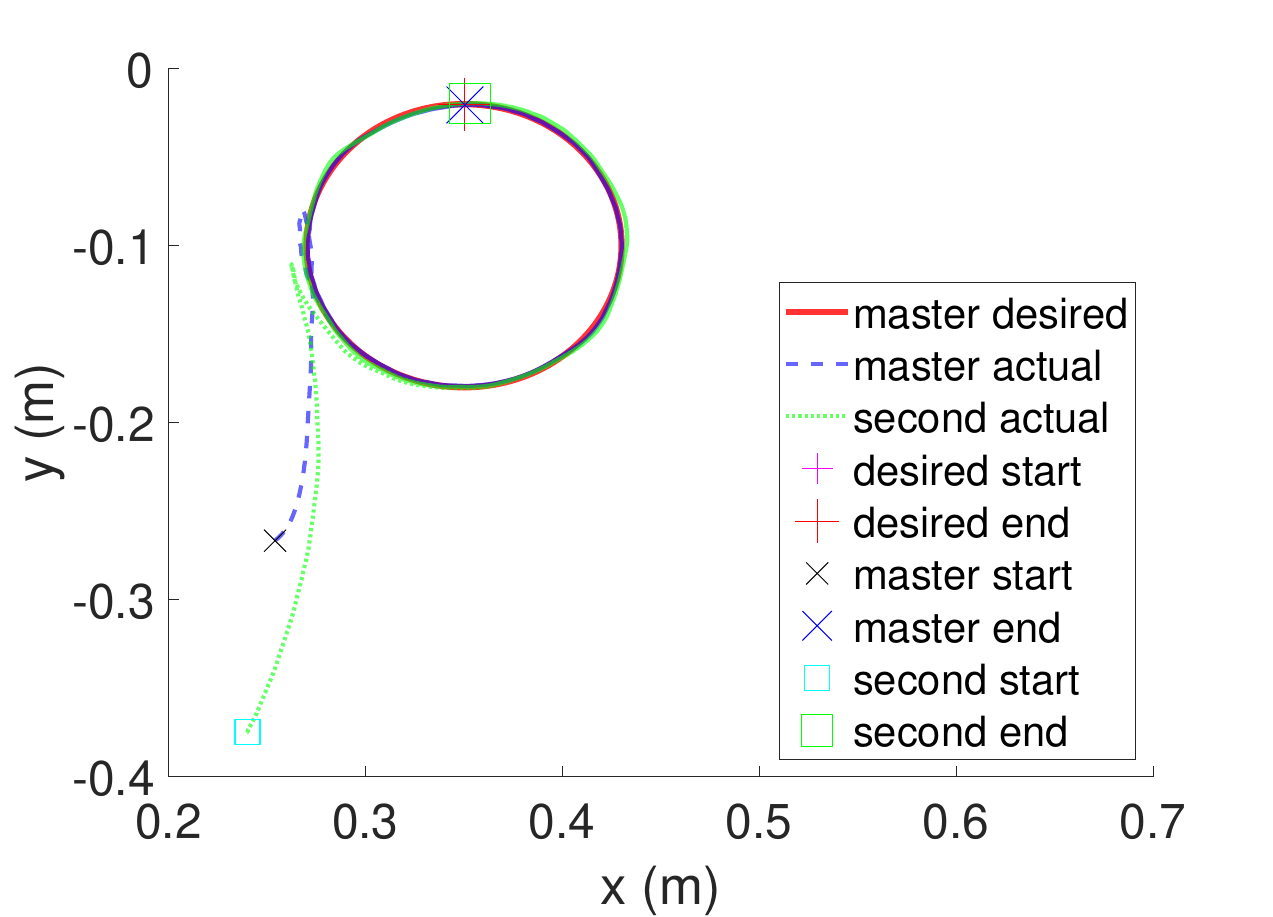}
         \caption{Circle trajectory (\#1)}
         \label{fig06a_trajectory_exp_traj123}
     \end{subfigure}
     \hfill
     \begin{subfigure}[b]{0.24\textwidth}
         \centering
         \includegraphics[width=\textwidth]{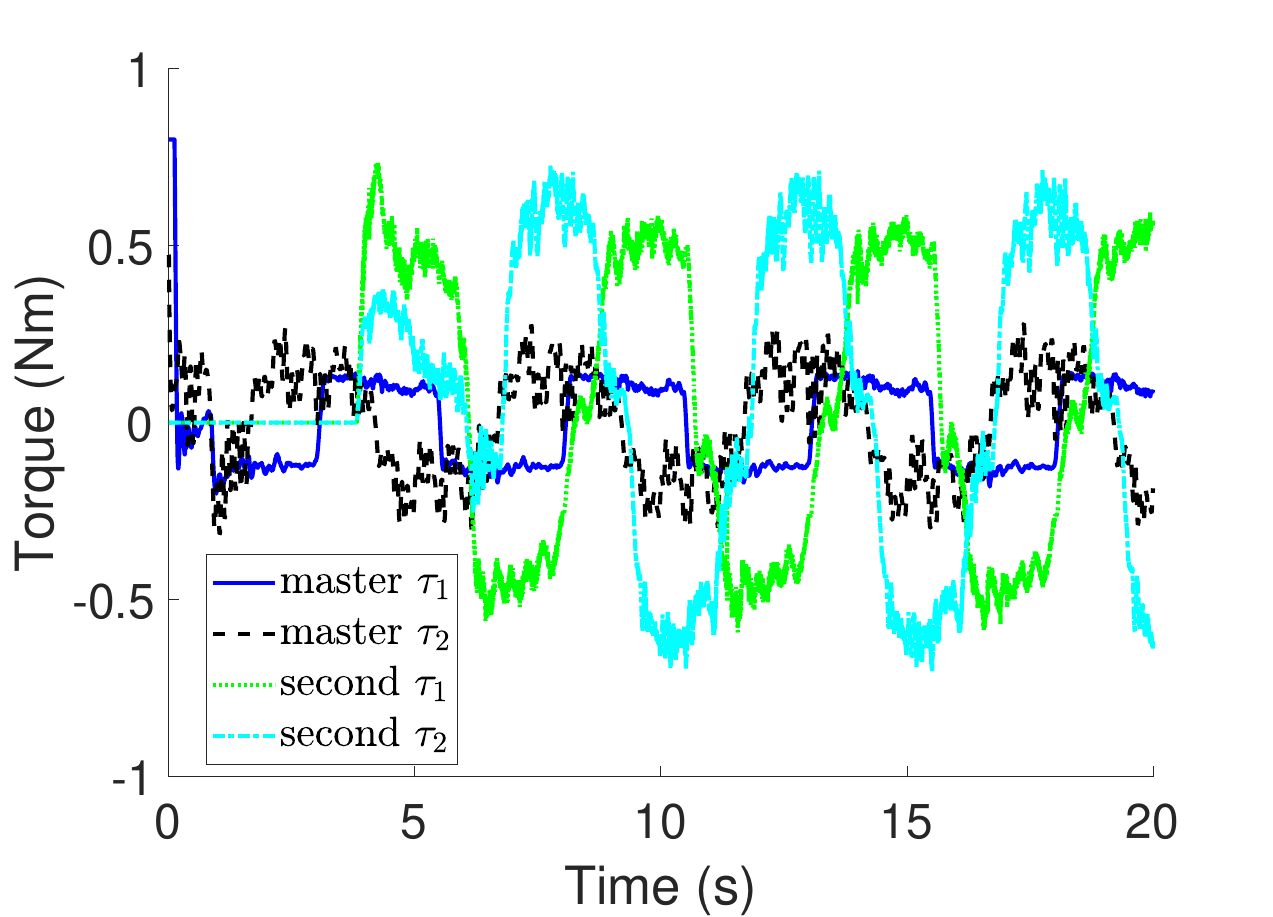}
         \caption{Torque of trajectory \#1}
         \label{fig06b_trajectory_exp_traj123}
     \end{subfigure}
     \hfill
     \begin{subfigure}[b]{0.24\textwidth}
         \centering
         \includegraphics[width=\textwidth]{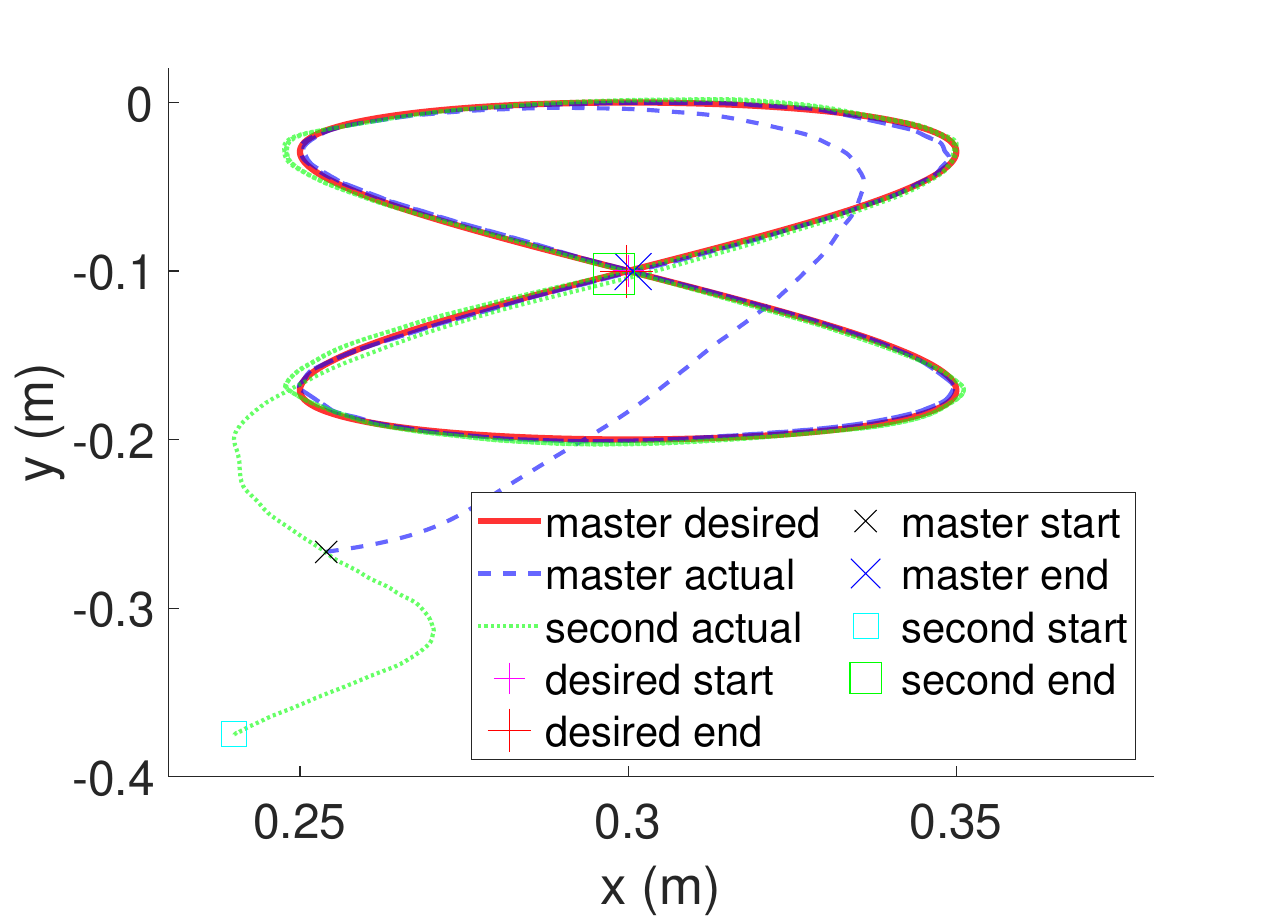}
         \caption{Figure-eight trajectory (\#2)}
         \label{fig06c_trajectory_exp_traj123}
     \end{subfigure}
     \hfill
     \begin{subfigure}[b]{0.24\textwidth}
         \centering
         \includegraphics[width=\textwidth]{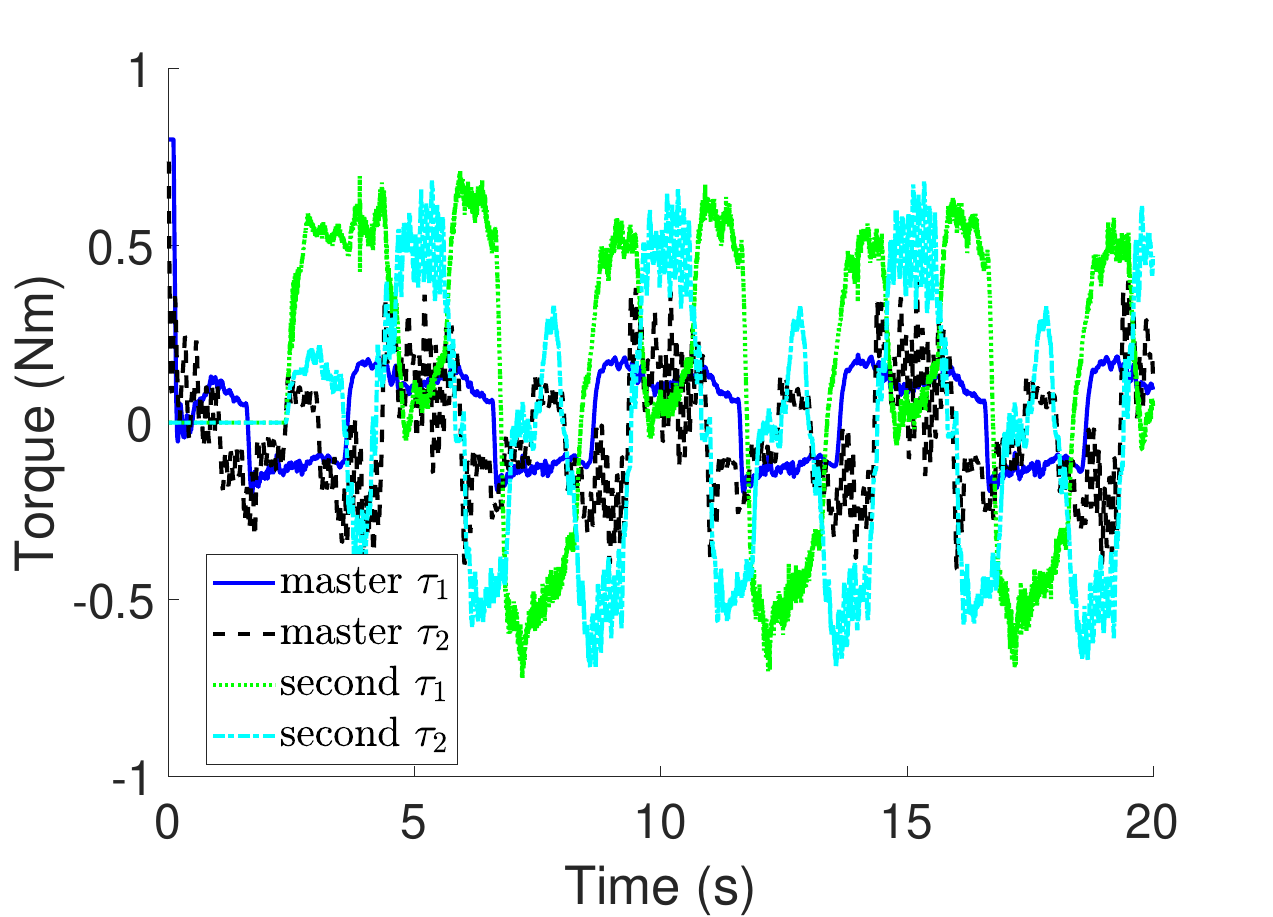}
         \caption{Torque of trajectory \#2}
         \label{fig06d_trajectory_exp_traj123}
     \end{subfigure}
     \hfill
     \begin{subfigure}[b]{0.24\textwidth}
         \centering
         \includegraphics[width=\textwidth]{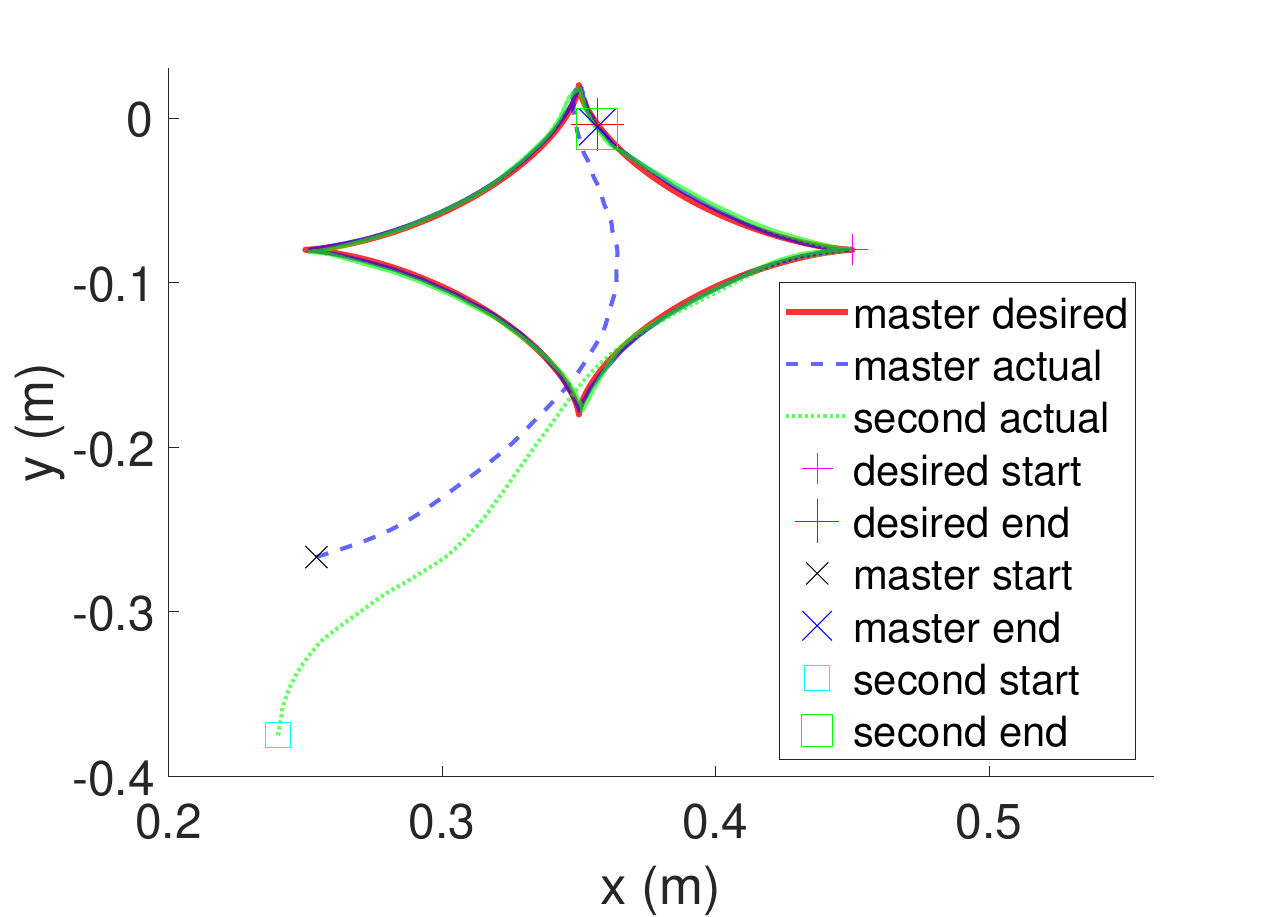}
         \caption{Tetragon trajectory (\#3)}
         \label{fig06e_trajectory_exp_traj123}
     \end{subfigure}
     \hfill
     \begin{subfigure}[b]{0.24\textwidth}
         \centering
         \includegraphics[width=\textwidth]{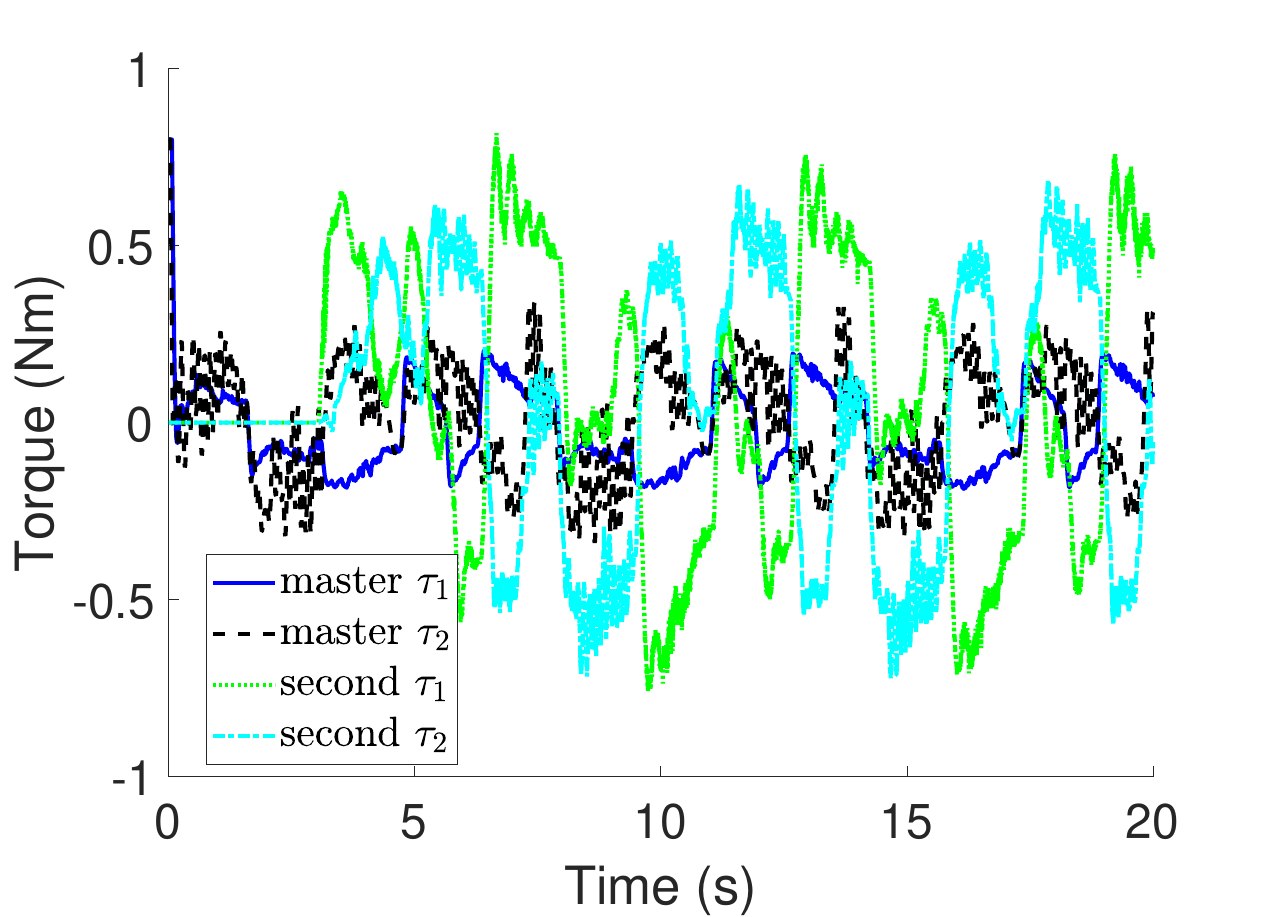}
         \caption{Torque of trajectory \#3}
         \label{fig06f_trajectory_exp_traj123}
     \end{subfigure}
     \hfill
     \begin{subfigure}[b]{0.24\textwidth}
         \centering
         \includegraphics[width=\textwidth]{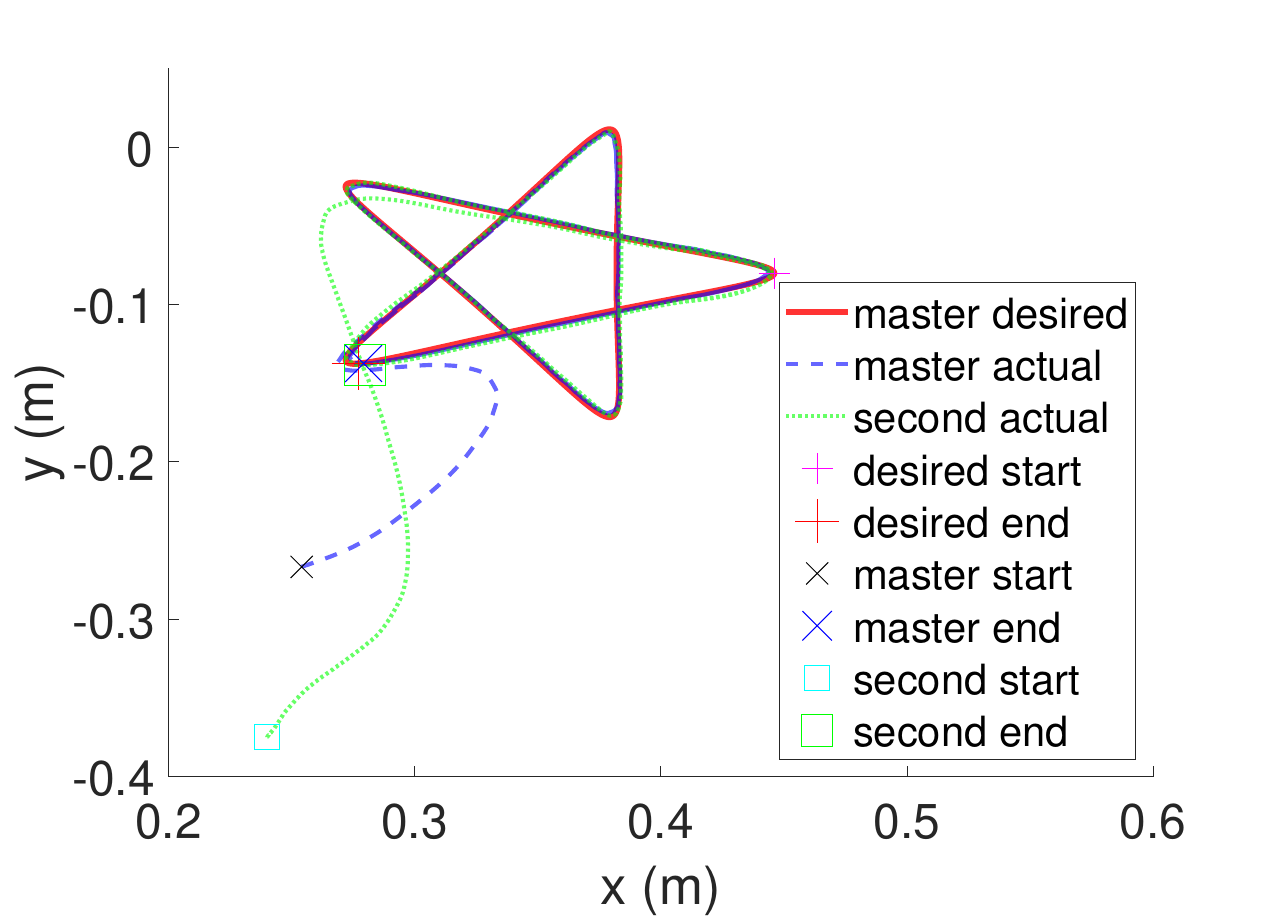}
         \caption{Pentagram trajectory (\#4)}
         \label{fig06g_trajectory_exp_traj123}
     \end{subfigure}
     \hfill
     \begin{subfigure}[b]{0.24\textwidth}
         \centering
         \includegraphics[width=\textwidth]{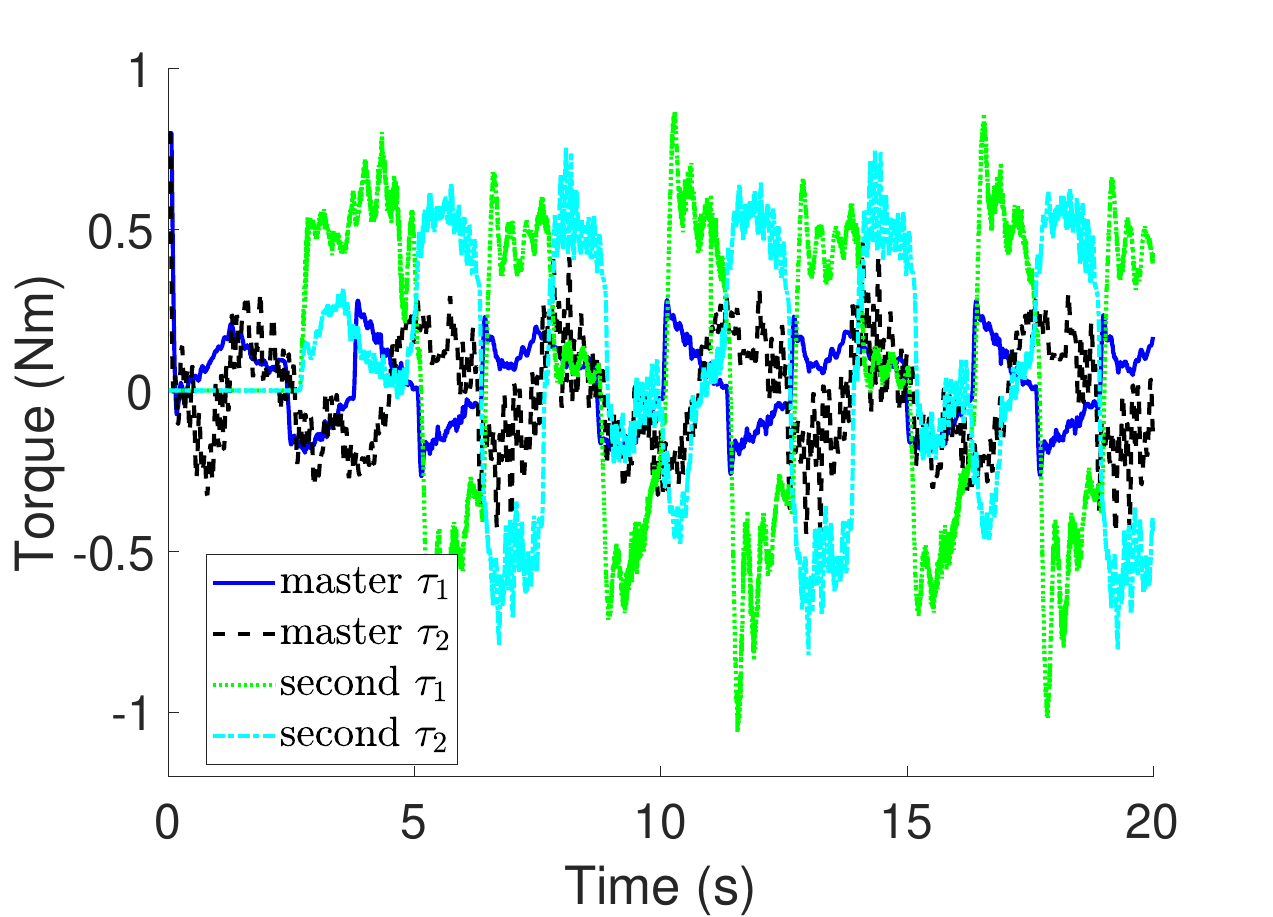}
         \caption{Torque of trajectory \#4}
         \label{fig06h_trajectory_exp_traj123}
     \end{subfigure}
     \hfill
     \begin{subfigure}[b]{0.24\textwidth}
         \centering
         \includegraphics[width=\textwidth]{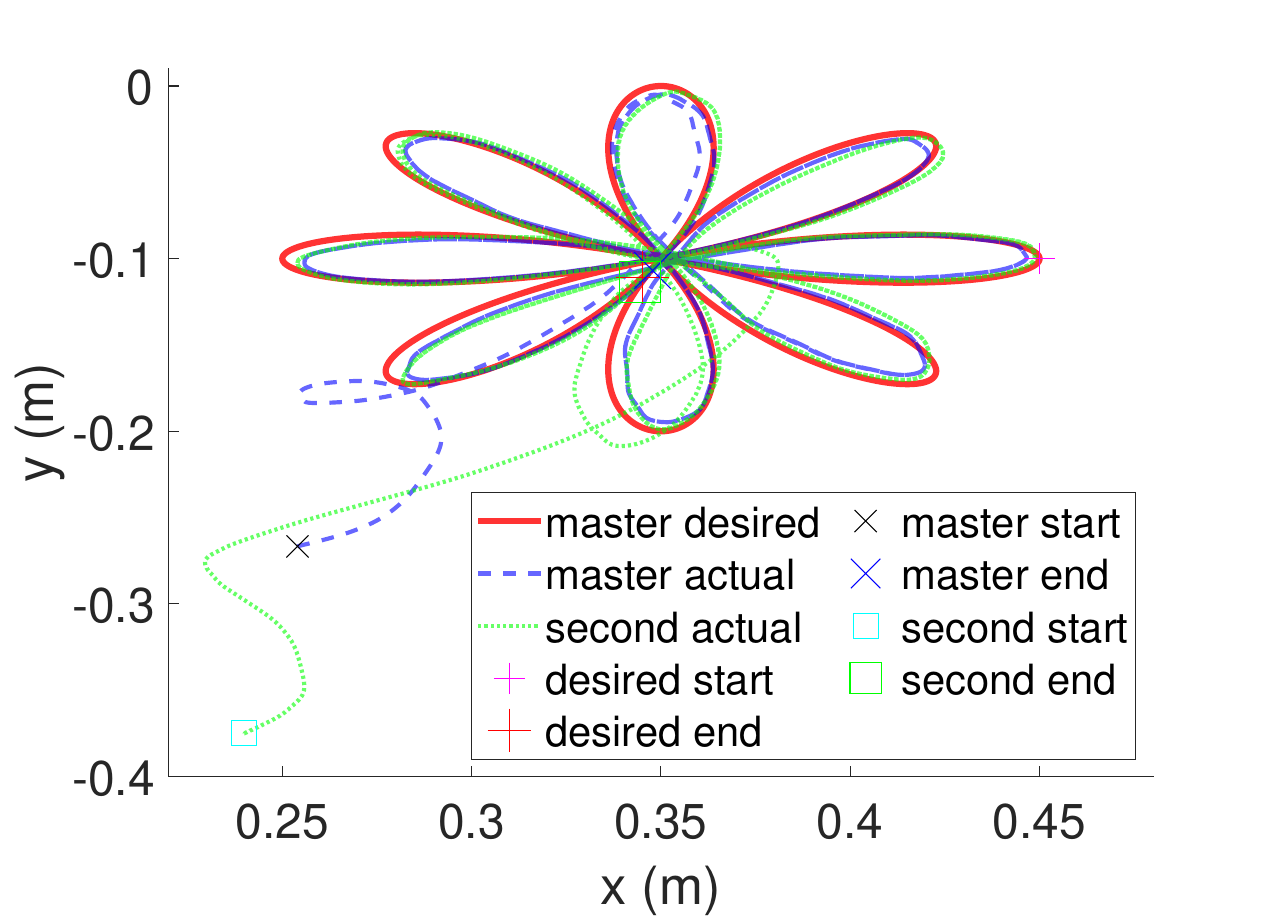}
         \caption{Rose trajectory (\#5)}
         \label{fig06i_trajectory_exp_traj123}
     \end{subfigure}
     \hfill
     \begin{subfigure}[b]{0.24\textwidth}
         \centering
         \includegraphics[width=\textwidth]{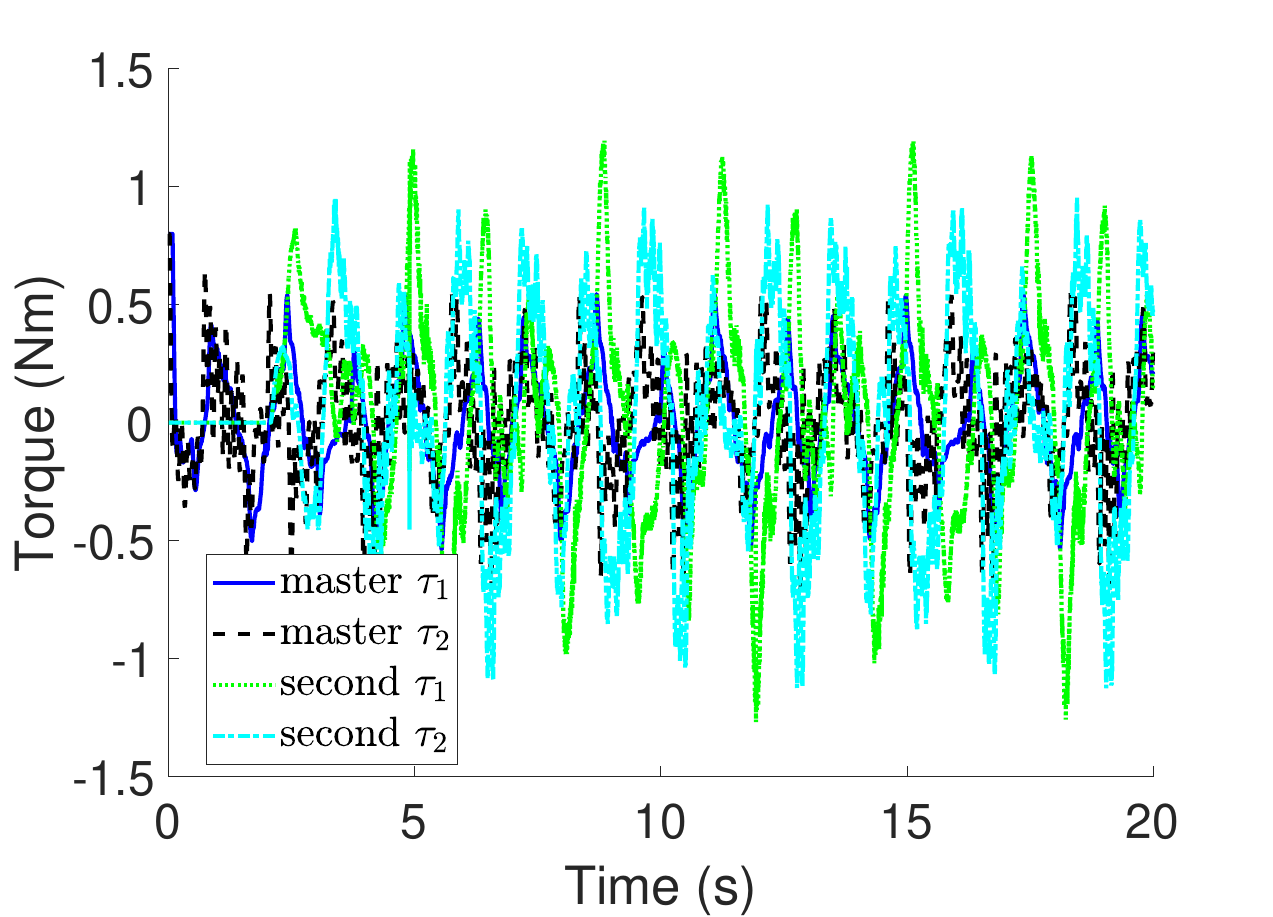}
         \caption{Torque of trajectory \#5}
         \label{fig06j_trajectory_exp_traj123}
     \end{subfigure}
        \caption{Exp.2 results of the teleoperation system performance on trajectory tracking tasks. The master robot tracks a pre-defined trajectory while the second robot follows the master robot. }
        \label{fig06_trajectory_exp_traj123}
\end{figure}

Fig.~\ref{fig06_trajectory_exp_traj123} shows the tracking performance of the teleoperation system. As can be seen in the figure, the teleoperation system can accurately perform the trajectory tracking tasks from a simple pattern to a complex pattern (Fig.~\ref{fig06a_trajectory_exp_traj123}, \ref{fig06c_trajectory_exp_traj123}, \ref{fig06e_trajectory_exp_traj123}, \ref{fig06g_trajectory_exp_traj123}, \ref{fig06i_trajectory_exp_traj123}). Although the tracking accuracy of the complex rose trajectory pattern (Fig.~\ref{fig06i_trajectory_exp_traj123}) is not as good as the simple ones (Fig.~\ref{fig06a_trajectory_exp_traj123}, \ref{fig06c_trajectory_exp_traj123}, \ref{fig06e_trajectory_exp_traj123}, \ref{fig06g_trajectory_exp_traj123}), it can be further improved by fine-tuning the impedance gains and observer gains. From the figure, we can also notice that the actual torques for the two robots are not on the same level, and the second robot needs larger torques (maximum around $0.5$ Nm) than the master robot (maximum around $0.2$ Nm) for tracking the same trajectory. This is reasonable since they are not identical robots, although they have the same form of kinematics. This also reflects that the NDOB can accurately estimate the dynamic uncertainties for the second robot (white) although it employs an identified dynamic mode that is not actually for it.

The results of Exp.2 indicate that the teleoperation system can accurately perform trajectory tracking tasks with varying pre-defined patterns by implementing an impedance controller and NDOB observer. The master robot (black) can accurately follow the pre-defined trajectory while the second robot (white) can accurately follow the master robot.

\subsection{Exp.3: pHRI}

In actual robot-assisted rehabilitation scenarios, pre-defined trajectories might not be always suitable for patients due to their various rehabilitation requirements. Then, the trajectory needs to be customized for individual patients by the therapist which can more effectively help the patient to recovery. In Exp.3, physical human-robot interaction ($p$HRI) is involved for the master robot where the therapist moves the master robot EE manually in order to customize trajectories while the second robot follows the trajectory of the master EE. 

In this experiment, for the master robot, only an impedance controller is implemented since an NDOB observer will prevent human-robot interaction \cite{teng4IROS2022NDOB}. The configuration of the second robot remains unchanged, \ie{}, both impedance controller and NDOB are implemented. The teleoperation control in Exp.3 is bilateral, \ie{}, the master robot will provide force feedback to its operator when the second robot is in contact with the surrounding environment. Even though, in Exp.3, the operator of the master robot will not receive force feedback since there is no interaction between the second robot and the surrounding environment.

\begin{figure}
     \centering
     \begin{subfigure}[b]{0.24\textwidth}
         \centering
         \includegraphics[width=\textwidth]{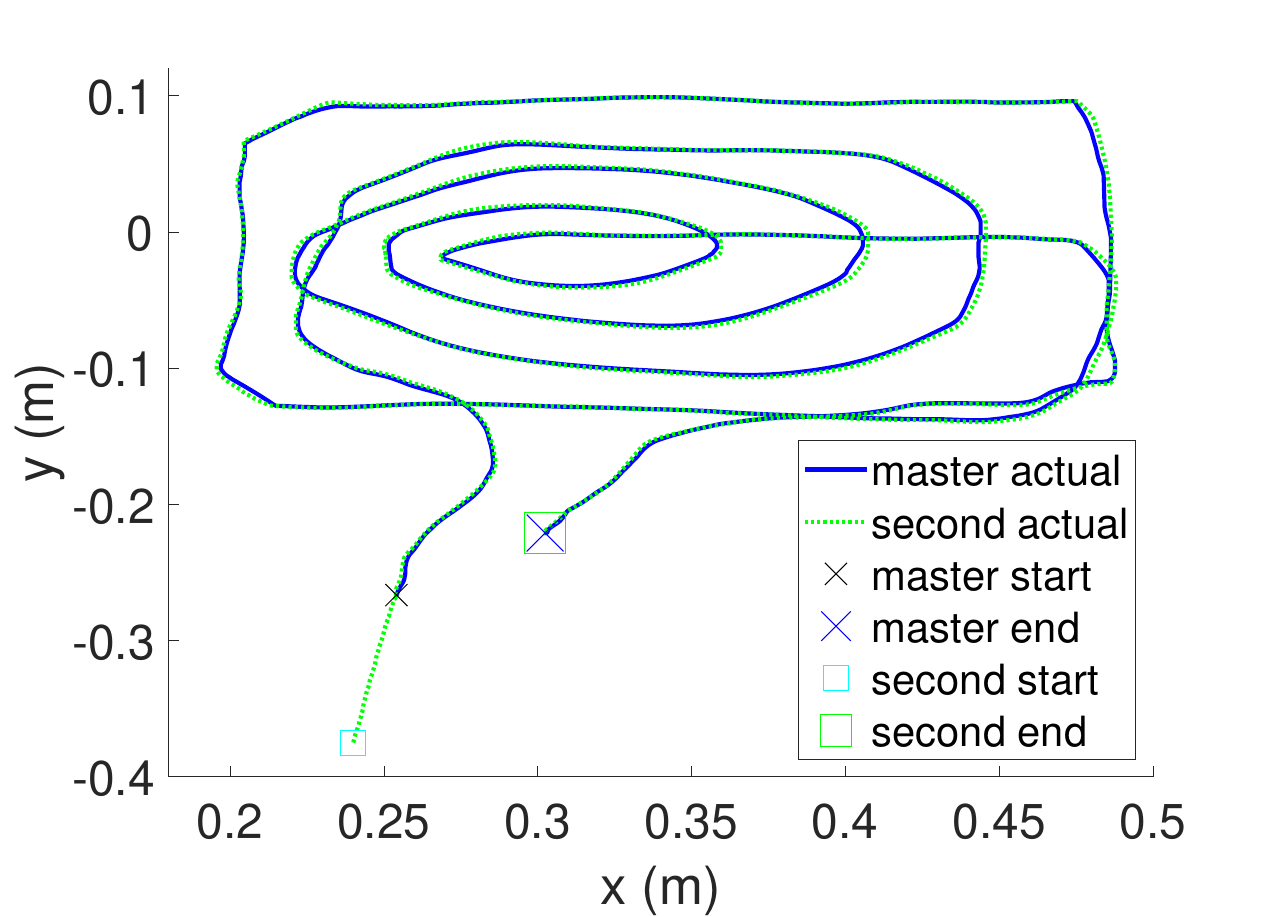}
         \caption{Customized trajectory}
         \label{fig06a_trajectory_exp06_pHRI}
     \end{subfigure}
     \hfill
     \begin{subfigure}[b]{0.24\textwidth}
         \centering
         \includegraphics[width=\textwidth]{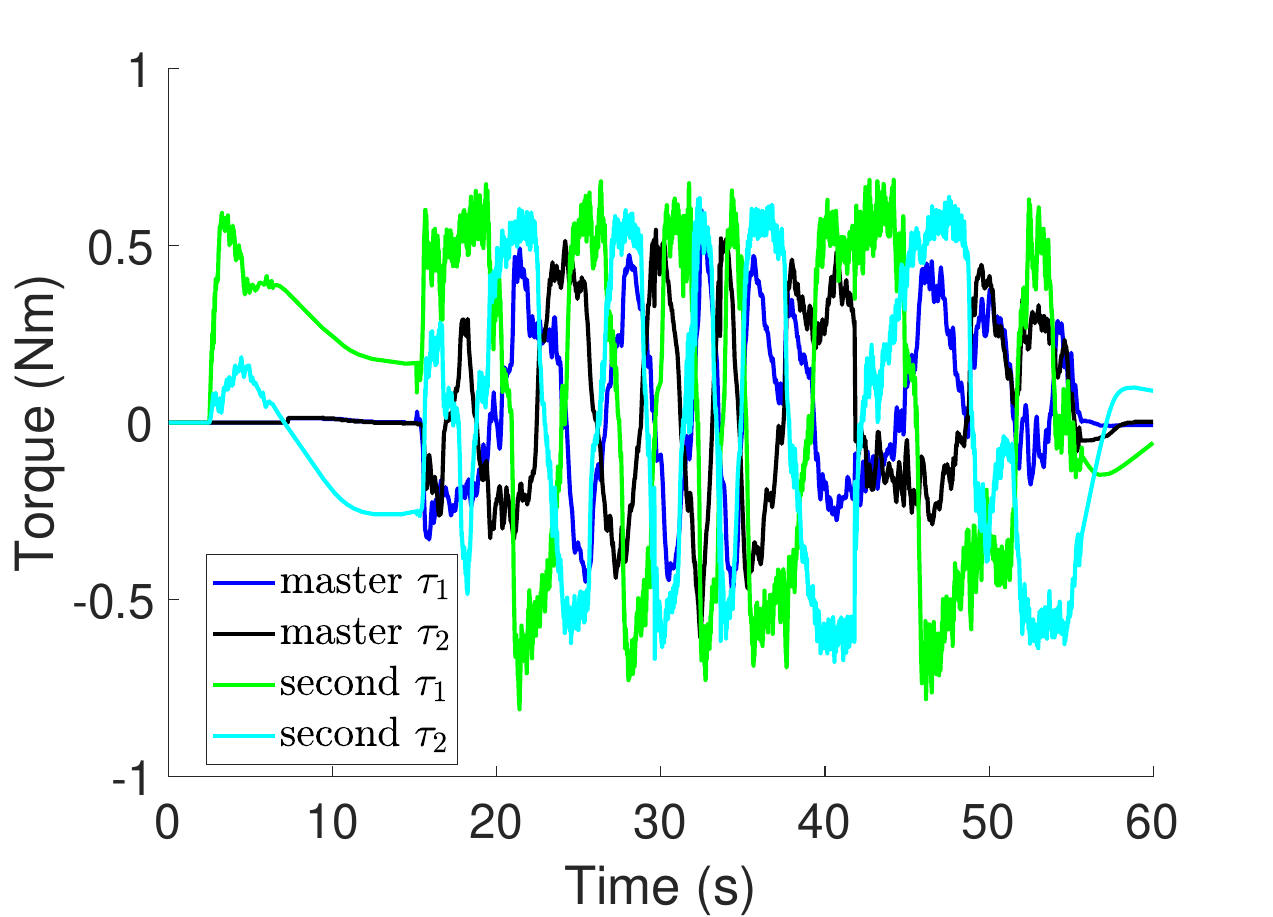}
         \caption{Actual torque}
         \label{fig06b_trajectory_exp06_pHRI}
     \end{subfigure} 
        \caption{Exp.3 results when the master robot is in $p$HRI mode. The trajectory of the master robot is defined by the operator while the second robot follows the master robot. }
        \label{fig06_trajectory_exp06_pHRI}
\end{figure}

Fig.~\ref{fig06_trajectory_exp06_pHRI} shows the experimental results when the master robot is operated manually by the operator in the $p$HRI mode while the second robot follows the trajectory of the master robot EE. The results show that the operator of the master robot can customize the trajectory for the master robot, while the second robot can accurately follow the master robot.

\subsection{Exp.4: Force Feedback}

In Exp.3, $p$HRI mode, the operator of the master robot does not receive rendered force feedback due to the fact that the second robot is in free motion and has no interaction with the surrounding environment. In Exp.4, force feedback will be evaluated when the second robot is in contact with external objects. Two scenarios are designed. In scenario 1, the second robot will be in contact with an external stiff wall while the master robot will render force feedback accordingly and provide it to its operator. In scenario 2, the second robot EE is attached to an external object ($1130$ grams) and drags it around mimicking a rehabilitation scenario of the patient's limb being attached to the second robot EE and passively moving around for rehabilitation training.

\begin{figure}
     \centering
     \begin{subfigure}[b]{0.24\textwidth}
         \centering
         \includegraphics[width=\textwidth]{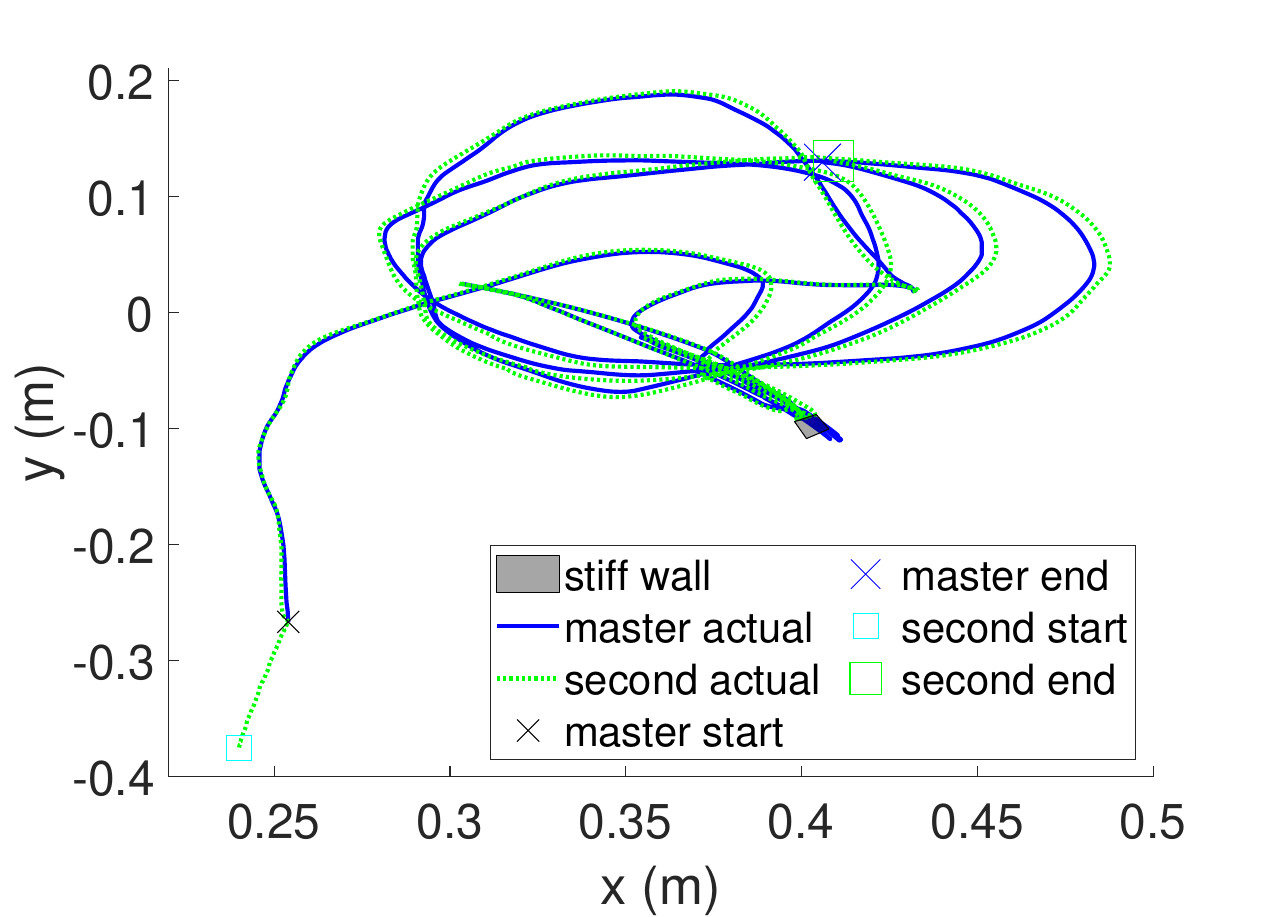}
         \caption{Trajectory}
         \label{fig07a_trajectory_exp07_pHRI_Fslave}
     \end{subfigure}
     \hfill
     \begin{subfigure}[b]{0.24\textwidth}
         \centering
         \includegraphics[width=\textwidth]{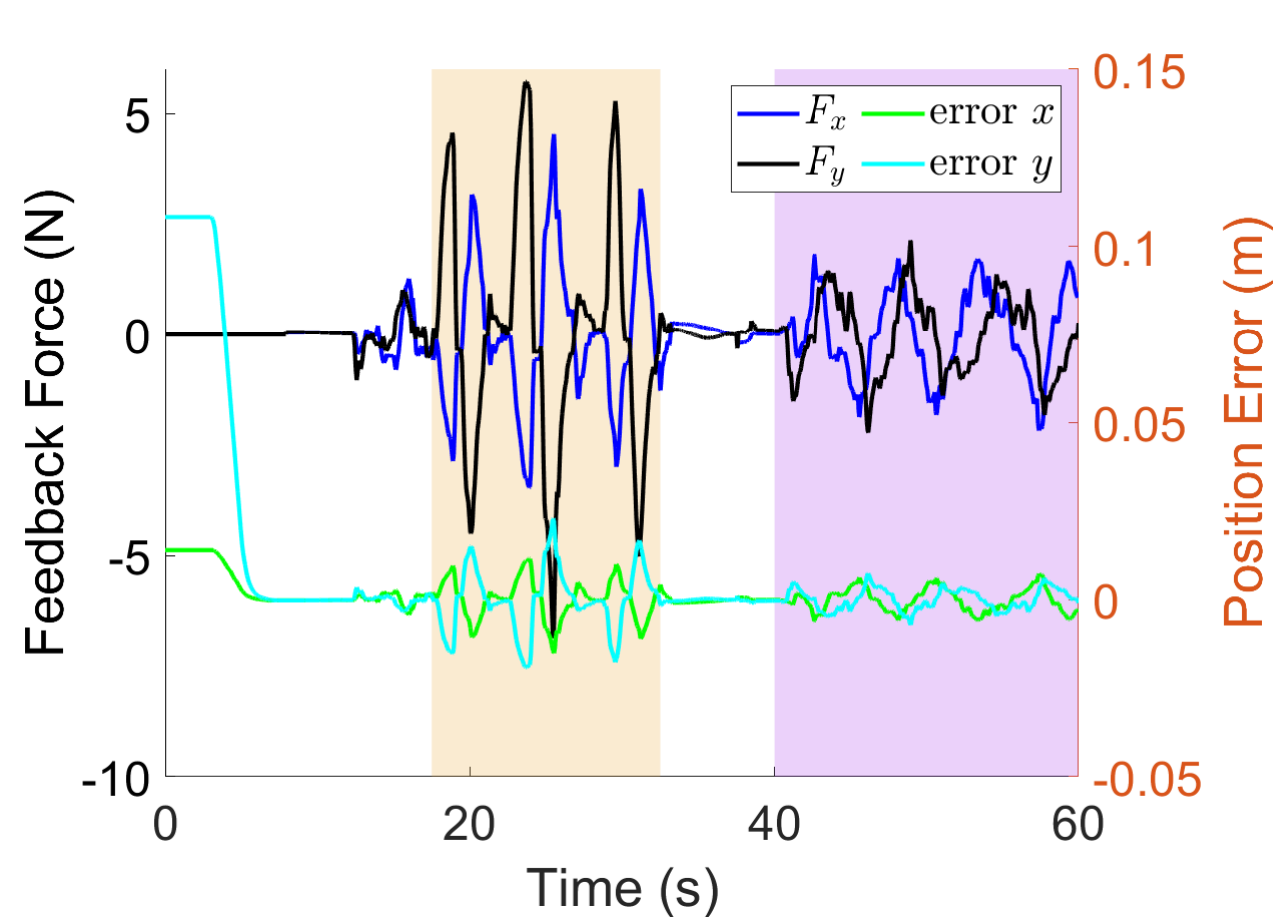}
         \caption{Force Feedback}
         \label{fig07b_trajectory_exp07_pHRI_Fslave}
     \end{subfigure}
     \hfill
     \begin{subfigure}[b]{0.24\textwidth}
         \centering
         \includegraphics[width=\textwidth]{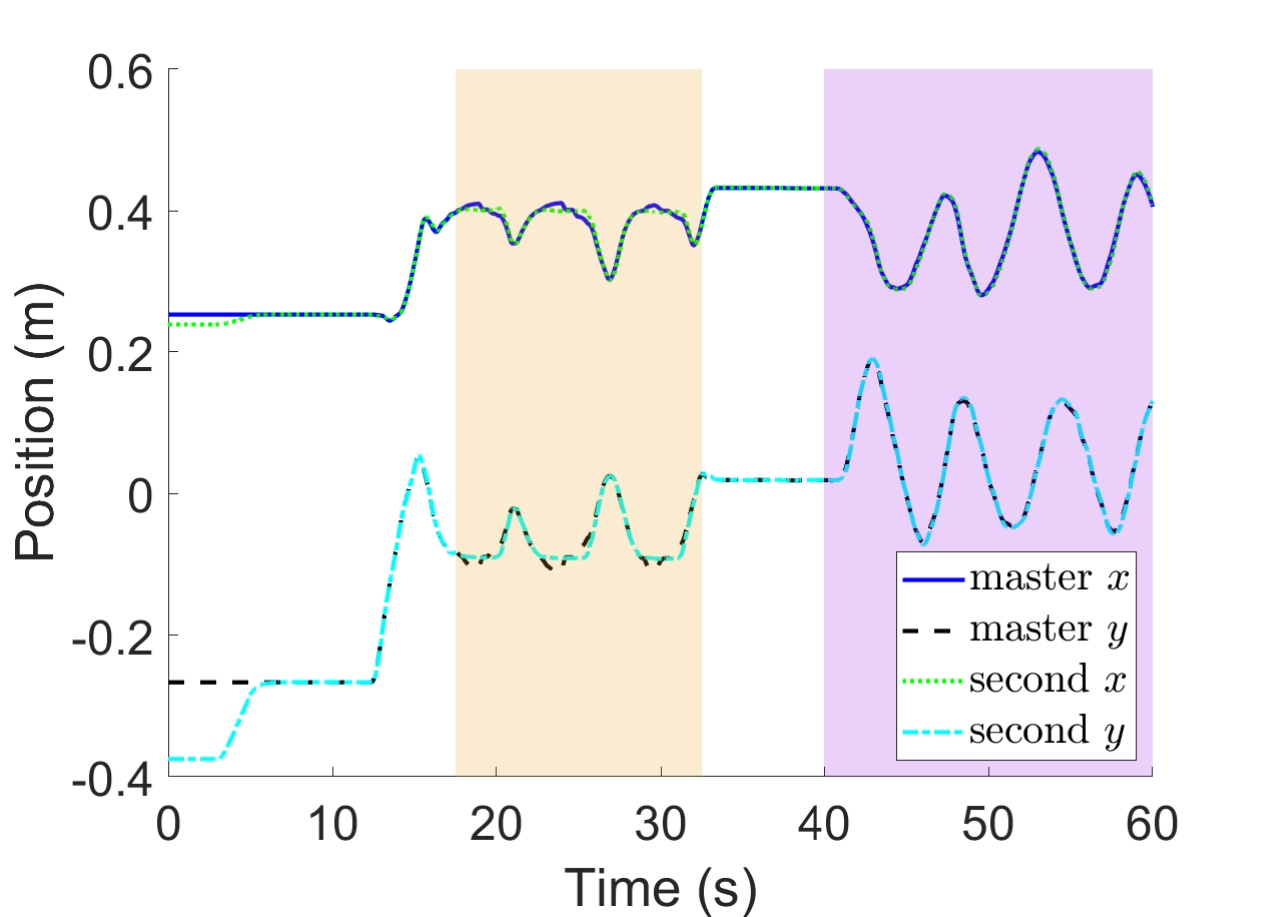}
         \caption{Position}
         \label{fig07c_trajectory_exp07_pHRI_Fslave}
     \end{subfigure}
     \hfill
     \begin{subfigure}[b]{0.24\textwidth}
         \centering
         \includegraphics[width=\textwidth]{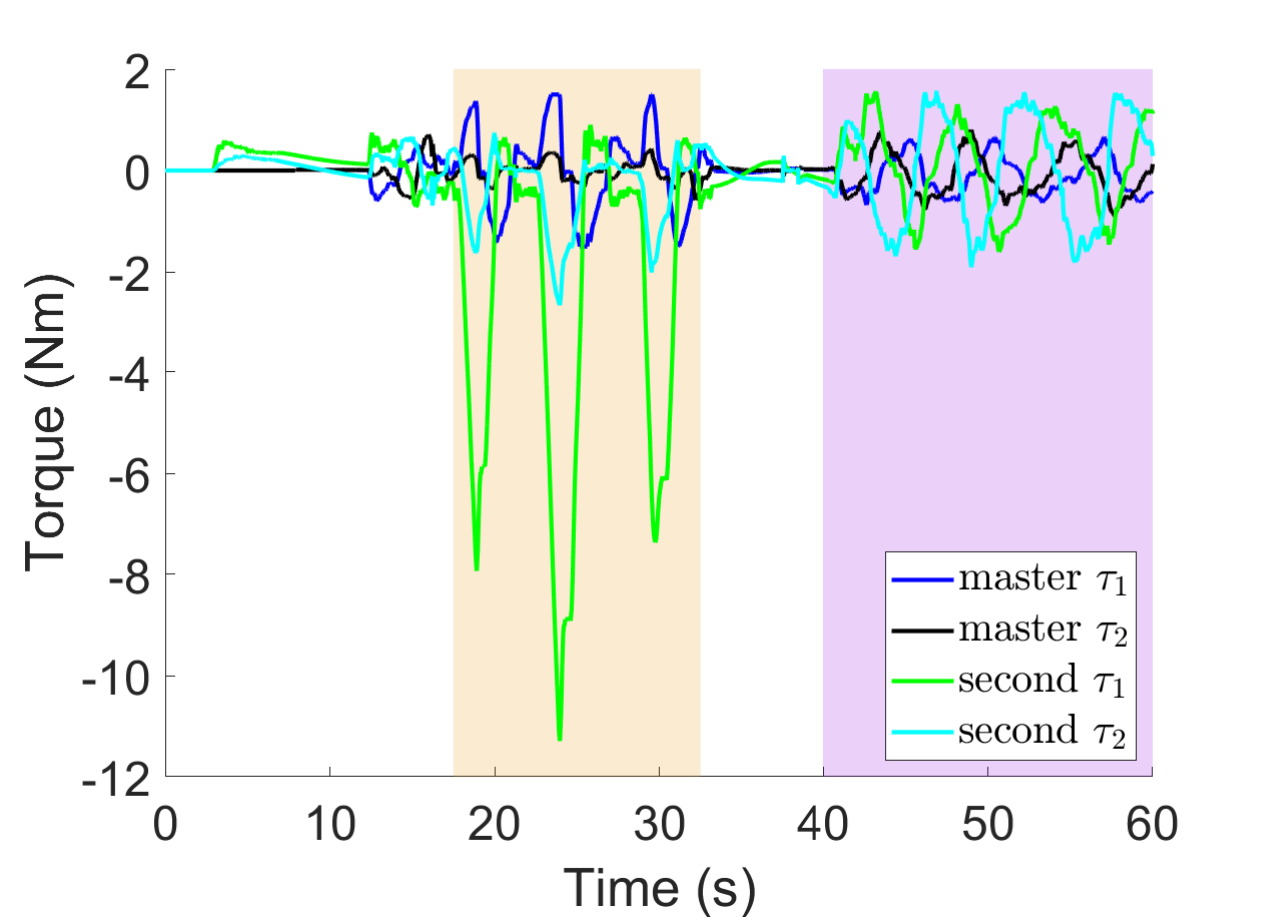}
         \caption{Actual torque}
         \label{fig07d_trajectory_exp07_pHRI_Fslave}
     \end{subfigure}
        \caption{Exp.4 results of force feedback rendering on the master robot side when the second robot interacts with external objects. The yellow-colored area represents scenario 1, and the purple-colored area represents scenario 2. }
        \label{fig07_trajectory_exp07_pHRI_Fslave}
\end{figure}

The experimental results of Exp.4 are shown in Fig.~\ref{fig07_trajectory_exp07_pHRI_Fslave}, where the yellow-colored area represents scenario 1, and the purple-colored area represents scenario 2. In scenario 1 of the yellow-colored area, the operator remotely controls the second robot to probe a stiff wall by operating the master robot while the contact force between the second robot and the stiff wall is rendered by the master robot and delivered to its operator. As can be seen in Fig.~\ref{fig07b_trajectory_exp07_pHRI_Fslave}, the rendered force feedback is in the range of $[-7, 7]$ N. In scenario 2 of the purple-colored area, the operator remotely controls the second robot to drag the patient's upper limb (represented by an external object of $1130$ g), while the operator receives force feedback which is in the range of $[-3, 3]$ N. Note that in this work, we only consider scenarios where the patients move their limbs passively, thus we represent the patient's limb by an external object.

The Exp.4 results show that the feedback force can be appropriately rendered according to the position error between the master robot EE and the second robot EE, and then delivered to the operator of the master robot. The feedback force here can be rendered to be stiffer or softer by tuning the corresponding gain.

\subsection{Exp.5: Record-Replay}

In conventional rehabilitation, the therapist often needs to help the patient move the limb repetitively along a customized trajectory, which is time-consuming and strenuous. This can be easily and effectively done by the proposed teleoperation system. The therapist can demonstrate a customized trajectory only once which can be recorded by the robot, then the robot can repeat the trajectory as many times as needed on its own.

\begin{figure}
     \centering
     \begin{subfigure}[b]{0.24\textwidth}
         \centering
         \includegraphics[width=\textwidth]{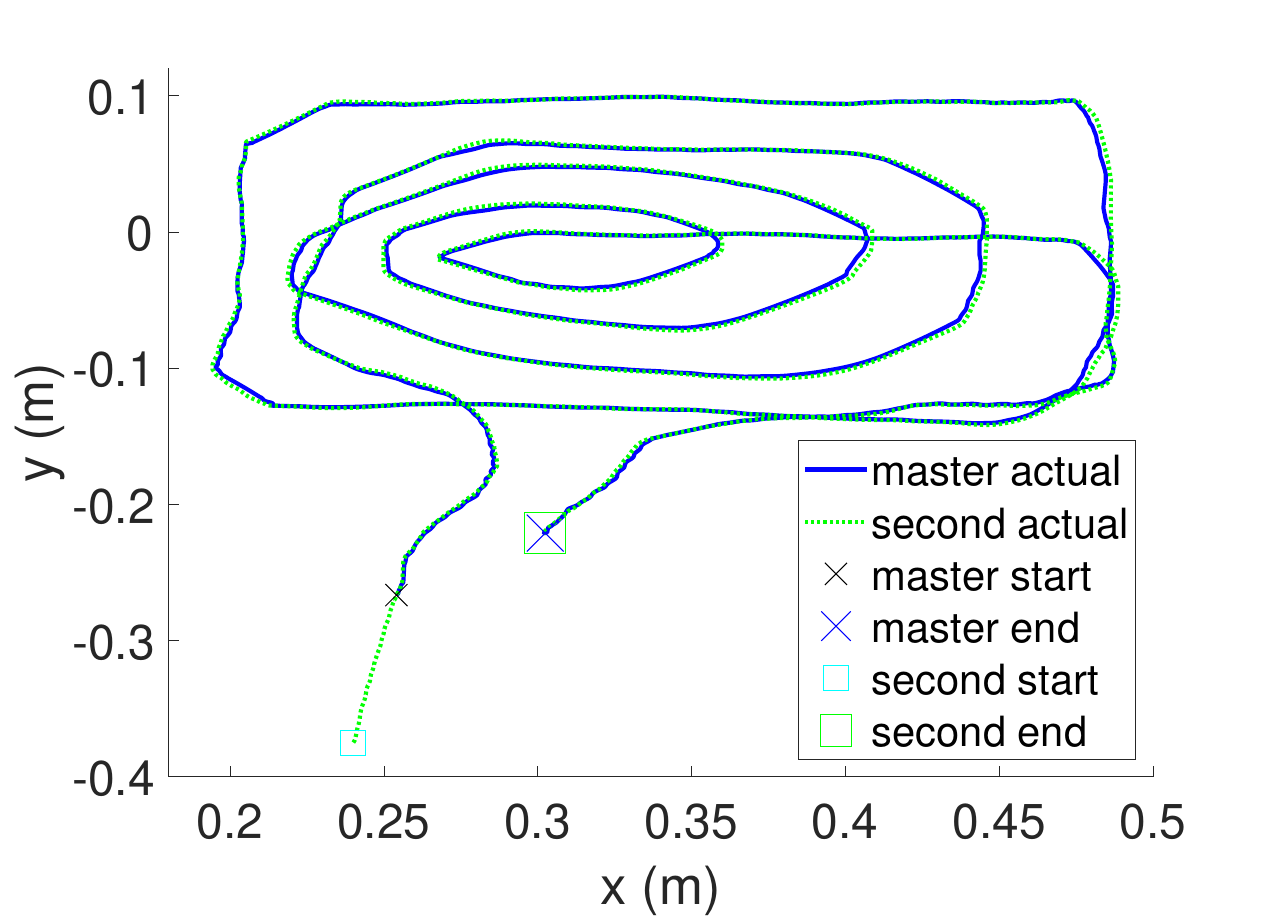}
         \caption{Trajectory}
         \label{fig08a_trajectory_exp08_pHRI_replayExp06}
     \end{subfigure}
     \hfill
     \begin{subfigure}[b]{0.24\textwidth}
         \centering
         \includegraphics[width=\textwidth]{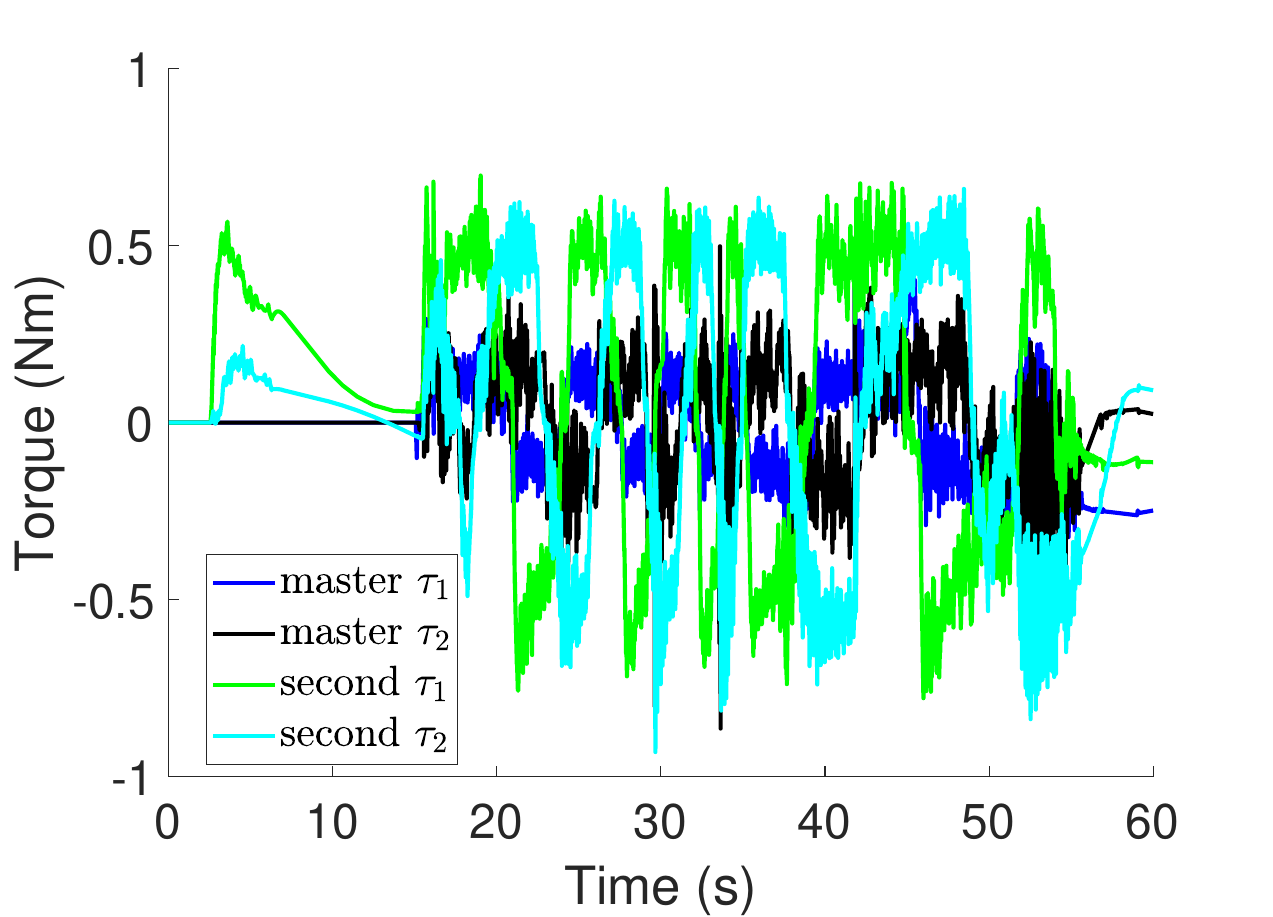}
         \caption{Actual torque}
         \label{fig08b_trajectory_exp08_pHRI_replayExp06}
     \end{subfigure} 
        \caption{ Exp.5 results of replaying the recorded trajectory data from Exp.3. The master robot replays the recorded trajectory while the second robot follows the master robot. }
        \label{fig08_trajectory_exp08_pHRI_replayExp06}
\end{figure}

In previous Exp.3, the customized trajectory demonstrated by the operator is recorded. Here in Exp.5, we let the master robot replay the recorded trajectory on its own. The experiment results are shown in Fig.~\ref{fig08_trajectory_exp08_pHRI_replayExp06}. As 
can be seen in the figure, the teleoperation system can accurately replay the recorded trajectory by comparing it with its original demonstration (Fig.~\ref{fig06_trajectory_exp06_pHRI}). Note that the actual torque (Fig.~\ref{fig08b_trajectory_exp08_pHRI_replayExp06}) is a bit noisy than that in the original demonstration (Fig.~\ref{fig06b_trajectory_exp06_pHRI}). This is reasonable since the recorded velocity and acceleration are discontinuous numerical values. As long as the sampling frequency of the recorded trajectory is sufficiently high, \eg{}, 1000 Hz in this experiment, the noise in the actual torque can be largely minimized and has no adverse effect on the system performance.

\section{Discussions}
\label{discussions}

In this paper, we propose a teleoperation system for robot-assisted rehabilitation training where patients can passively move their upper limbs with the help of a robot. The control scheme integrated and enabled three main features in one teleoperation system. First, the proposed system allows the therapist to program pre-defined trajectories with various patterns using their mathematical expressions (trajectory-tracking mode), thus helping the patient conduct movement therapy training repetitively. Second, the proposed system allows the therapist to operate the master robot ($p$HRI mode), then the second robot on the patient side will follow the master robot accurately. In such a way, the therapist can directly help the patient to conduct the movement therapy training remotely. Third, the system can record the customized movement of the therapist in $p$HRI mode, then replay it repetitively on its own in trajectory-tracking mode to help the patient do the training.

In the proposed teleoperation system, the master robot (on the therapist side) is implemented with both an impedance controller and a disturbance observer in trajectory-tracking mode, while with only an impedance controller in $p$HRI mode. The second robot (on the patient side) is implemented with both an impedance controller and a disturbance observer in both trajectory-tracking mode and $p$HRI mode since it is designed to always follow the master robot. For the master robot in $p$HRI mode, the disturbance observer needs to be turned off since it will prevent human-robot interaction \cite{teng4IROS2022NDOB}.

For the master robot in $p$HRI mode, force feedback is rendered and delivered to the operator (the therapist) when the second robot is in contact with the surrounding environment. This allows the therapist to track the real-time status of the patient via the force feedback. The force feedback is rendered based on the position error between the master robot EE and the second robot EE rather than the actual interaction force. The advantage is that, by appropriate gain-tuning, the rendered force can be tuned to be stiffer or softer, and it is independent of the controller. 

In \cite{tavakoli2007highFidelitybilateralTeleop}, Tavakoli \etal{} summarized typical teleoperation architectures with force feedback including two-channel (2CH) and 4CH control architectures. The 2CH position-error-based (PEB, also called position-position) architecture is a symmetric design where the master robot takes the real-time position of the second robot as its desired position, and vice versa. This 2CH PEB architecture does not need a force sensor while force feedback that is proportional to the position difference will be rendered automatically on both sides due to the symmetric design. However, this design suffers from a distorted perception even when the robot is in free motion due to the non-perfect position tracking (\ie{}, non-zero position errors), which means that the operator of the master robot will receive proportional force feedback even when the second robot is not in contact with the surrounding environment. Another 2CH architecture is direct force reflection (DFR, also called force-position). This architecture requires a force sensor to measure the interaction force between the second robot and the environment, then only the measured force (not including the position anymore) is sent back to the master robot for generating force feedback. Although the 2CH DFR is proven to be better than the 2CH PEB in terms of force feedback transparency, the operator can still feel the intrinsic inertia of the master robot when the second robot is not in contact with any objects. The 4CH architecture is able to achieve ideal transparency by installing force sensors on both sides to measure the interaction forces. Both the measured force and the real-time position will be sent to the other robot for generating force feedback and position tracking. However, installing force sensors makes the system complex, and usually unnecessary. When measuring the interaction force is not needed on the master (or second) robot side, the corresponding force sensor can be removed, thus a 4CH architecture can be reduced to a 3CH architecture without imposing additional expense on the system transparency. This makes the 3CH architecture extremely attractive. Furthermore, all these 2CH/3CH/4CH architectures are designed for $p$HRI scenarios and they cannot be directly used for pre-defined trajectory tracking tasks.

In contrast to the 2CH/3CH/4CH architectures introduced in \cite{tavakoli2007highFidelitybilateralTeleop}, in this work, accurate force feedback (high transparency) is not necessary for rehabilitation. Therefore, force sensors are not needed. Instead, only the position error is used for generating a virtual spring force as the feedback on the master robot side, and by proper gain-tuning, the rendered force can be close to the actual interaction force. The proposed architecture is similar to the 2CH PEB architecture, \ie{}, both robot real-time positions will be sent to the other robot. But in our designed architecture, the master robot will only use the position of the second robot for force rendering. This also makes the force rendering be independent of the controller design. 

One advantage of the proposed architecture is that the teleoperation system can seamlessly switch from trajectory-tracking mode to $p$HRI mode while the architectures in \cite{tavakoli2007highFidelitybilateralTeleop} are devotedly focusing on $p$HRI mode. Another advantage is that the proposed control scheme is applicable to all general robotic systems rather than only haptic devices. On the contrary, a potential disadvantage is that the damper term in the controller (\ref{eqn_imped_controller_simplify4_pHRI_ff}) in $p$HRI mode may not be necessary for a backdrivable haptic device. For a haptic device, with the damper term, the operator will feel an additional damper force generated by this term, while without the damper term, the operator only feels the intrinsic inertia of the robot mechanism. If this is the case, the user can simply remove the damper term when using a haptic device.

A potential limitation of this work is that we are only considering the scenarios where the patients move their upper limbs passively. In future work, we will extend the current work to other scenarios where the patients can move their limbs actively to some extent, and then a robot can help the patients do the rehabilitation training cooperatively.

\section{Conclusions}
\label{conclusions}

In this paper, a bilateral teleoperation system is constructed for robot-assisted rehabilitation. The teleoperation system is implemented with a combination of impedance controller and disturbance observer where the former can provide compliant robot behavior for ensuring safe robot-human interaction while the latter can compensate for dynamic uncertainties when only a rough dynamic model is available. More importantly, the designed control architecture integrated both trajectory-tracking mode and $p$HRI mode, and allows them to be switched from one mode to another freely. The proposed teleoperation system demonstrated accurate trajectory tracking performance of both the master and second robot, as well as appropriate force feedback on the master robot side. Moreover, the teleoperation system allows the therapist to demonstrate a customized trajectory only once, and then the system, on its own, can accurately replay the recorded trajectory as many times as needed.

The proposed teleoperation system is promising to be used for robot-assisted rehabilitation. It can reduce the workload of the therapist and retain the quality of the work of helping the patient move along a pre-defined or customized trajectory for rehabilitation training.

\section*{Acknowledgment}

The author would like to thank Dr. Mahdi Tavakoli and Dr. Milad Nazarahari for their invaluable and insightful feedback and suggestions on improving the quality of this work.

\ifCLASSOPTIONcaptionsoff
  \newpage
\fi

\bibliographystyle{IEEEtran}
\bibliography{IEEE_bib}

\end{document}